\documentclass[11pt]{article}
\usepackage[a4paper, total={6in, 8in}]{geometry}

\usepackage{graphicx}%
\usepackage{multirow}%
\usepackage{amsmath,amssymb,amsfonts}%
\usepackage{amsthm}%
\usepackage{mathrsfs}%
\usepackage[title]{appendix}%
\usepackage{xcolor}%
\usepackage{textcomp}%
\usepackage{manyfoot}%
\usepackage{booktabs}%
\usepackage{algorithmicx}%
\usepackage{algpseudocode}%
\usepackage{listings}%

\usepackage{hyperref}

\usepackage{amsmath,amssymb,amsfonts}
\usepackage{graphicx}
\usepackage{textcomp}

\usepackage{subfig}

\usepackage{algpseudocode}
\usepackage{algorithm2e}

\usepackage{amsmath}
\usepackage{array}

\usepackage{amssymb}
\usepackage{graphicx}

\usepackage{url}

\usepackage{pifont}

\newcolumntype{C}[1]{>{\centering\let\newline\\\arraybackslash\hspace{0pt}}m{#1}}

\usepackage{placeins}

\usepackage{natbib}[sort]

\begin{document}

\date{}
\title{Reasoning in machine vision by learning fast and slow thinking}

\maketitle

\begin{center}
{\large{Shaheer U. Saeed$^{1,2,3,4,5\ast}$, Yipei Wang$^{4,5}$, \\
Veeru Kasivisvanathan$^{6,7}$, Brian R. Davidson$^{8}$,\\
Matthew J. Clarkson$^{4,5}$, Yipeng Hu$^{4,5,9}$, Daniel C. Alexander$^{4,9}$}}
\end{center}

\begin{center}
{\normalsize{$^{1}$Centre for Bioengineering, Queen Mary University of London, London, UK} \\
\normalsize{$^{2}$Digital Environment Research Institute, Queen Mary University of London, London, UK} \\
\normalsize{$^{3}$School of Engineering and Materials Science, Queen Mary University of London, London, UK} \\
\normalsize{$^{4}$UCL Hawkes Institute, University College London, London, UK}\\
\normalsize{$^{5}$Department of Medical Physics and Biomedical Engineering, University College London, London, UK}\\
\normalsize{$^{6}$Centre for Urology Imaging, Prostate, AI and Surgical Studies (COMPASS) Research Group, Division of Surgery and Interventional Science, University College London, London, UK} \\
\normalsize{$^{7}$Department of Urology, Comprehensive Cancer Center, Medical University of Vienna, Vienna, Austria}\\
\normalsize{$^{8}$Division of Surgery and Interventional Science, University College London, London, UK} \\
\normalsize{$^{9}$Department of Computer Science, University College London, London, UK} \\
\normalsize{$^\ast$Correspondence e-mail: shaheer.saeed@qmul.ac.uk}
}    
\end{center}

\abstract{
Reasoning is a hallmark of human intelligence, enabling adaptive decision-making in complex unfamiliar scenarios. In contrast, machine intelligence remains bound to training data, unable to dynamically refine solutions at inference. While recent advances have explored machine reasoning - trading inference-time compute for improved performance - they focus on verbal domains such as mathematical problem-solving where explicit rules govern step-by-step solution generation. Many tasks lack sufficient labelled data and require alternative performance improvement mechanisms, such as inference-time compute. Here we present a paradigm for machine reasoning in vision, enabling performance improvements with increasing thinking time (inference-time compute), even with limited labelled data. Our approach is inspired by dual-process theories of human cognition, integrating a fast-thinking System I module for generating and verifying solutions in familiar tasks, with a slow-thinking System II module that iteratively refines predictions using self-play reinforcement learning, even when task-specific data is limited. This paradigm involves proposing, competing over, and refining solutions until convergence. We demonstrate that extended inference-time compute yields superior performance compared to large-scale supervised learning, foundation models, and human experts in vision tasks. These include computer-vision benchmarks and cancer localisation across five organs, highlighting the potential of inference-time compute for data-scarce problems.}

\section*{Introduction}

Rethinking machine cognition - beyond data-driven learning:
Human cognition excels at new tasks with minimal prior experience through reasoning, by dedicating cognitive resources, such as extended thinking time, to planning and strategising \cite{kahneman2003maps, evans1984heuristic, kahneman20122}. 
This is enabled by dual-process human cognition \cite{kahneman2003maps}: System I enables fast, intuitive decisions based on prior experience, while System II facilitates slow, deliberate reasoning, especially in unfamiliar scenarios.
In the context of machines, reasoning or System II thinking is analogous to the ability to consume computational resources, such as inference-time compute, to improve performance, as opposed to using data for improvement, as noted in recent literature \cite{anthony2017thinking, deepseek_r1, wei2022chain}.  
Conventional deep learning systems lack this capacity to perform reasoning or System II thinking \cite{anthony2017thinking, evans2003two}, relying heavily on extensive human-labelled data or repeated task exposure to perform well in familiar scenarios \cite{litjens2017survey}. While such data-driven experience yields strong performance within the scope of training data, these systems falter when labelled data is insufficient to represent new tasks \cite{litjens2017survey, bhattacharya2022review, koh2022artificial}. Here, we propose an algorithm which enables System II thinking and equips machines with the ability to improve performance with increasing thinking time (inference-time computation), rather than resorting to additional labelled data.

Two systems of human cognition and their transition in psychology:
Theories from cognitive psychology have observed that fast and intuitive System I thinking
allows humans to make confident decisions for familiar tasks, despite unclear immediate justifications~\cite{kahneman2003maps, evans1984heuristic, kahneman20122}. 
In contrast, System II thinking consumes cognitive resources at the time of the decision-making \cite{kahneman2003maps}. For instance, a beginner chess player can reason about potential moves, plan and develop complex strategies using only the game rules, despite limited gameplay experience. 
One key cognitive resource consumed in such processes is thinking time and its mechanisms include comparison of hypothetical scenarios, considering long-term outcomes in greater detail, and breaking problems into simpler parts before solving them \cite{evans2003two, kahneman2003maps}. Attention is another critical cognitive resource \cite{kahneman2003maps}, which can be consumed through mechanisms such as filtering out irrelevant information (e.g., ignoring background speech while reading) \cite{sorqvist2015concentration}. Consuming these resources allows improvements in decision quality, which is an ability considered unique to humans \cite{evans2003two}.

With experience, humans can transition tasks from slow, deliberate System II to much faster System I thinking \cite{kahneman20122} e.g., the ability of experienced chess players to execute strategies more quickly and instinctively as their experience grows. The interplay between fast and slow thinking allows versatility in human cognition, which is capable of handling both familiar and unfamiliar scenarios.

\begin{figure}
    \centering
    \includegraphics[width=0.95\linewidth]{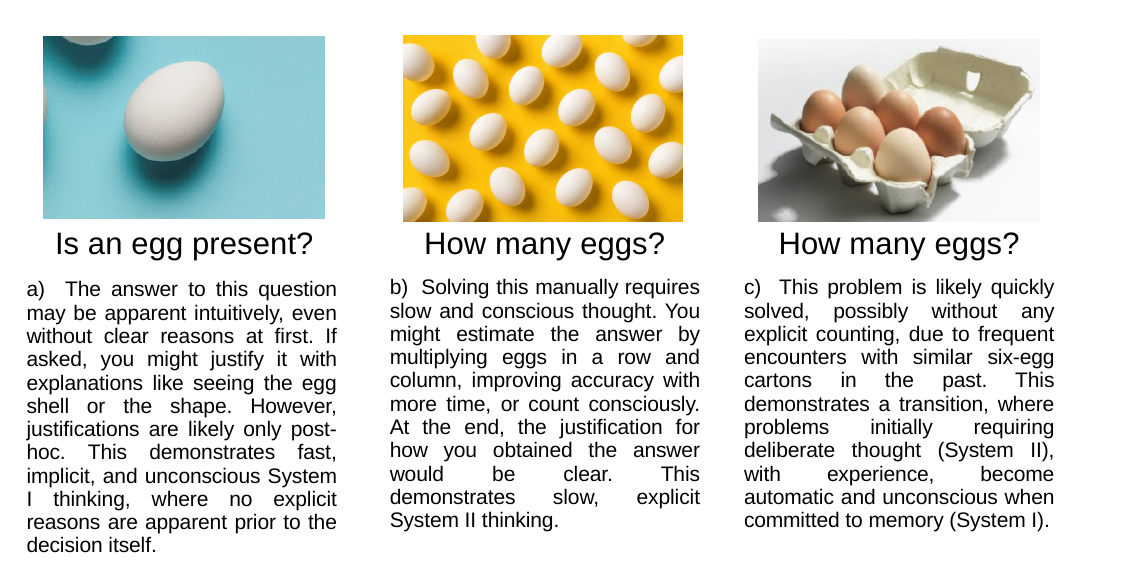}
    \caption{Examples illustrating System I and System II cognitive processes. a) and b) represent problems solved by System I and System II thinking, respectively. c) represents a problem that has transitioned from System II to System I over time. Note: a), b) and c) are open stock photos available via an institutional Microsoft 365 subscription.}
    \label{fig:combined_eggs}
\end{figure}

We present examples, in Fig. \ref{fig:combined_eggs}, showcasing tasks typically solved by System I and System II thinking, along with tasks that have likely transitioned between the two.

Shortfalls of machine cognition:
Using the analogy of dual-process cognition, machines are limited in the sense that their predictions are largely feed forward and made with fixed inference times, i.e., they operate only in a System I regime. In other words, supervised deep learning algorithms are highly dependent on large labelled data sets and lack mechanisms to improve by extending thinking time (inference-time compute) \cite{koh2022artificial, anthony2017thinking, evans2003two}, or handle scenarios beyond the training distribution \cite{vabalas2019machine}.
Approaches like few-shot learning \cite{wang2020generalizing, ravi2016optimization, snell2017prototypical} and foundation models \cite{kirillov2023segment, ma2024segment, zhou2023foundation, moor2023foundation} aim to use task-specific data more efficiently. However, they exhibit fixed, often linear, computational complexity with respect to data availability and the only viable mechanism to improve performance further is to train with more human-labelled data \cite{mazurowski2023segment, xie2024sam, li2024adapting}. 

The analogue for System II capabilities in machines would allow performance improvement with an increase in computation time, overcoming practical challenges stemming from the cost and scarcity of labelled data, as noted in recent literature \cite{vabalas2019machine, wei2022chain, hosseini2024v, deepseek_r1,liu2025can, su2024dualformercontrollablefastslow, anthony2017thinking, you2024biochemicalprostatecancerrecurrence, ganapini2022combining, fabiano2023fast, pham2023continual, qi2024interactive}.

Progress towards inference-time compute capabilities in machines:
There has been prior work implementing inference-time compute strategies for machine inference, that are analogous to System II thinking. However, such approaches have largely focused on constrained and restrictive domains, such as board games \cite{anthony2017thinking}. These approaches modelled System I and System II separately, without explicit modelling of the connection between the two. In these works the System II slow-thinking process is implemented as an exhaustive, structured, random or guided search over promising moves, as a mechanism to consume compute time \cite{anthony2017thinking, guo2014deep, silver2016mastering}. While these approaches improve performance by extending compute time without requiring additional task-specific data, they are limited to scenarios with manageable search spaces or to specialised domains such as language \cite{snell2024scaling, hosseini2024v, feng2023alphazero, trinh2024solving, xin2024deepseek}. Recent work has also modelled System II thinking using a reinforcement learning process guided by a reward signal. However, these remain limited to the language domain, relying largely on verifiable tasks such as mathematics problems, using rule-based verification of final outcomes or intermediate steps, to generate rewards \cite{deepseek_r1, kumar2024training, wang2023math, liu2025can}, sometimes involving human observers \cite{lightman2023let, uesato2022solving}. 

Several existing approaches aim to refine initial visual predictions, marking a step toward inference-time compute. Structured-prediction methods such as CRF-based refinements \cite{zheng2015conditional} enhance spatial coherence through hand-crafted constraints and a fixed number of update steps.  
Diffusion-based iterative approaches \cite{ho2020denoising} also refine predictions but follow predetermined generative trajectories, with fixed steps, and typically rely on large-scale datasets for training. Other methods improve data efficiency by selecting or adapting pretrained features \cite{dvornik2020selecting}, or by refining internal representations through hierarchical modules \cite{wang2025hierarchical, jolicoeur2025less}, though their inference-time computation is fixed once trained. Although System II reasoning operates during inference, it differs fundamentally from test-time adaptation approaches. Test-time adaptation typically updates the parameters of the predictive model to better match the test distribution. In contrast, in our framework the System I networks remain fixed, while the System II module optimises a reinforcement learning policy that iteratively refines candidate segmentations. The optimisation therefore occurs in the solution space rather than through adaptation of the segmentation model parameters. Inference-time compute analogous to System II thinking would allow performance to improve with added unbounded computation/ thinking time, even when labelled data is scarce.

Another type of cognitive resource in machines is attention which may also be considered as a model for System II thinking, although has been explored only in the context of natural language applications \cite{weston2023system}. 
Techniques such as `divide-and-conquer' \cite{cormen2022introduction} and `chain-of-thought' \cite{wei2022chain} break down complex tasks into parts before solving them. These mechanisms enable neural networks to improve through a refinement of their inputs, either guided by a human observer \cite{wei2022chain} or through iterative guided refinements of their own inputs which also increases inference-time compute \cite{weston2023system}. However, they remain limited to sequential query-response domains, such as large language models trained/fine-tuned on paired question-and-answer chat-based data, owing to the ease of verifying correctness in such domains. Arguably, attention mechanisms common in language applications do not model System II thinking \cite{john1994irrelevant, vsuch2019neural, laakom2021learning, vaswani2017attention}, as they do not posses an ability to dynamically alter behaviours for new samples or tasks, with attention often learnt based on task-specific experience itself.

Most of this previous work has focused on emulating verbal reasoning, however, non-verbal or visual reasoning has potential to enhance various applications including robotics, medical diagnostics, and astronomy. Many real-world tasks require reasoning about spatial relationships, object boundaries, and patterns without relying on language-based logic. In medical imaging, for instance, clinicians identify and interpret complex visual features to diagnose diseases—an inherently non-verbal reasoning process. Developing systems capable of such non-verbal reasoning is essential for advancing machine intelligence beyond text-based tasks and into broader, high-impact applications.

The main hindrance in implementing reasoning or inference-time compute for vision tasks, such as identifying object boundaries, stems from the fact that solutions are often more complex with much higher-dimensional solution spaces.
The methods above do not extend naturally to explore such spaces, which often have nuanced solutions with multiple nearly-correct outcomes. This makes development of the verification-based rewards, challenging in such domains. For example, even in a simple real-world 2D image segmentation to localise objects, the solution space is astronomically large, far beyond the computational capacity for exhaustive search, or for manual design of rule-based verification strategies (as an example, assuming a $512\times512$ image with a single object, there are $\sum_{r=0}^{512\times 512} \frac{(512\times 512)!}{r!((512\times 512)-r)!} \gg 10^{1000}$ possibilities; whereas for chess, 16 pieces per side, each with 16 moves, there are $\frac{16!}{1!(16-1)!} \times \frac{16!}{1!(16-1)!} = 256$ possibilities per-move).
Some efforts allow improved performance as a result of slow thinking \cite{su2024dualformercontrollablefastslow, you2024biochemicalprostatecancerrecurrence} but operate with fixed thinking periods, limiting their scalability, with further performance improvements not possible with an increase in thinking time. Thus modelling of the reasoning enabled by System II thinking remains uncharted in the context of vision applications.

\begin{figure}[!ht]
    \centering
    \includegraphics[width=0.75\linewidth]{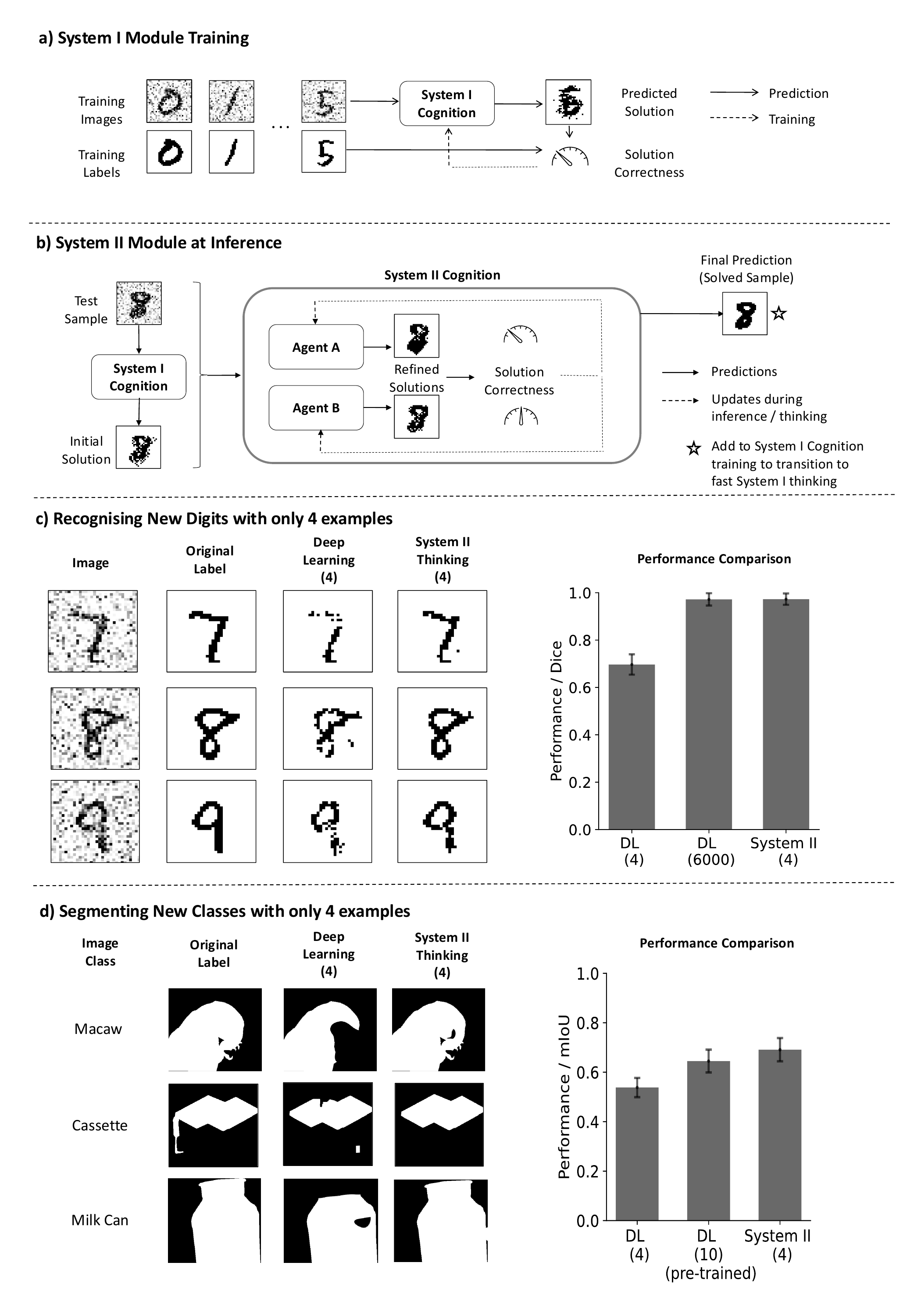}
    \caption{a) and b) present an overview of the training and evaluation of the proposed System II approach. c) and d) show performance of the System II approach compared with common methods (numbers of samples used for training/ adaptation, from the target task, in brackets). Note: images in a), b) and c) are derived from the public domain MNIST dataset, the original ImageNet images are omitted from d) due to copyright; however, these and corresponding segmentations are available via the online repository: \url{github.com/s-sd/system-II-vision} \cite{shaheer_u_saeed_2026_19331690}.}
    \label{fig:method_overview}
\end{figure}

An auto-competing mechanism to enable inference-time compute:
We propose two interconnected modules for enabling inference-time compute in machines, which are analogous to dual-process human cognition: 1) the System I module, trained on human-labelled data across various tasks, to produce fast solutions; and 2) the System II module, which allows for refinement and optimisation of solutions produced by the System I module. Even in data-limited scenarios, this framework allows performance improvements with added inference-time compute. Additionally, the System II module, after fully refining the solutions, feeds back to the System I module to improve its fast solutions for subsequent data, reducing time required for refinement in the System II module.
An overview of the described method is illustrated in Fig. \ref{fig:method_overview}a) and \ref{fig:method_overview}b). A detailed description is presented in the `Methods' section and a mathematical treatment is presented in the supplementary materials `methodological description'.

\FloatBarrier

\section*{Results}

\subsubsection*{Demonstration in standard computer vision tasks}

System II thinking recognises ``new'' digits with only four samples: Fig. \ref{fig:method_overview}a-c) illustrate how the System II method offers substantial improvement over traditional supervised learning in a simple vision task. For the task of segmenting digits from noisy images, we trained the System I module with digits 0-5 (as shown in Fig. \ref{fig:method_overview}a), and then adapted this using only 4 samples from each of the unseen target digits 6-9. The System II module was then used to segment 4000 samples from each of the target digits 6-9 (as shown in Fig. \ref{fig:method_overview}b). Fig. \ref{fig:method_overview}c), shows that our dual-process approach, using extended thinking time to improve performance, substantially outperformed a deep learning algorithm (trained separately from the System I module but following a similar training scheme) \cite{ronneberger2015u} trained and adapted with the same data. The System II approach, using only 4 samples from the target digits, also showed similar performance to a deep learning algorithm \cite{ronneberger2015u} trained with 6000 samples from the target digits, representing an upper-bound performance in these new tasks.

System II thinking improves segmentation on new image classes: Fig. \ref{fig:method_overview}d) illustrates the substantially improved segmentation performance, of the System II method, on unseen real-world image classes. We trained the System I module to segment objects using 2000 images and labels randomly sampled from the ImageNet-S dataset \cite{gao2022large}. We then adapted the System I module to segment 40 unseen target objects, using only 4 labelled samples for each. The System II module was then used to segment 6 unseen images for each of the 40 unseen objects. As shown in Fig. \ref{fig:method_overview}, System II thinking significantly outperformed the previous best deep-learning methods, not only when the deep learning methods used the same data, but also when they used substantially more pre-training data and labelled samples from the unseen target objects \cite{gao2022towards}. 

Supplementary materials section `detailed results for computer vision tasks' shows that improvements offered by the System II approach were statistically significant, for both experiments above.

\subsubsection*{Localisation of five types of abdomino-pelvic cancer}

Motivation: Cancer detection and localisation in medical images, such as prostate magnetic resonance (MR) and liver computed tomography (CT), holds significant promise for non-invasive diagnosis, often even outperforming the invasive alternatives \cite{ahmed2017diagnostic, krajewski2018imaging, oliva2004liver}. However, democratisation of these medical imaging technologies has been severely limited by the need for highly-specialised expertise and associated costs, often only accessible in a small number of large medical centres, even in the developed countries \cite{hunter2022role, koh2022artificial, sprague2016variation, ahmed2017diagnostic, vleugels2020suboptimal, czolbe2021segmentation}. Deep learning has been effective in segmenting anatomical structures, such as organs, where variability is relatively constrained \cite{ronneberger2015u, ma2024segment}. In contrast, cancer pathology presents categorically greater heterogeneity, with imaging manifestations that are less well-defined and continuously evolving due to new clinical discoveries and advancements in imaging technologies. This complexity is further compounded by the challenge of obtaining high-quality ground-truth labels, which require specialised clinical expertise or resource-intensive procedures such as independent histopathology examinations. Thus, cancer detection and localisation is a prime example where System I cognition falls short and System II cognition may offer data-efficient performance improvements that are comparable or even superior to what expert clinicians are capable of.

\begin{figure}[!ht]
    \centering
    \includegraphics[width=0.95\linewidth]{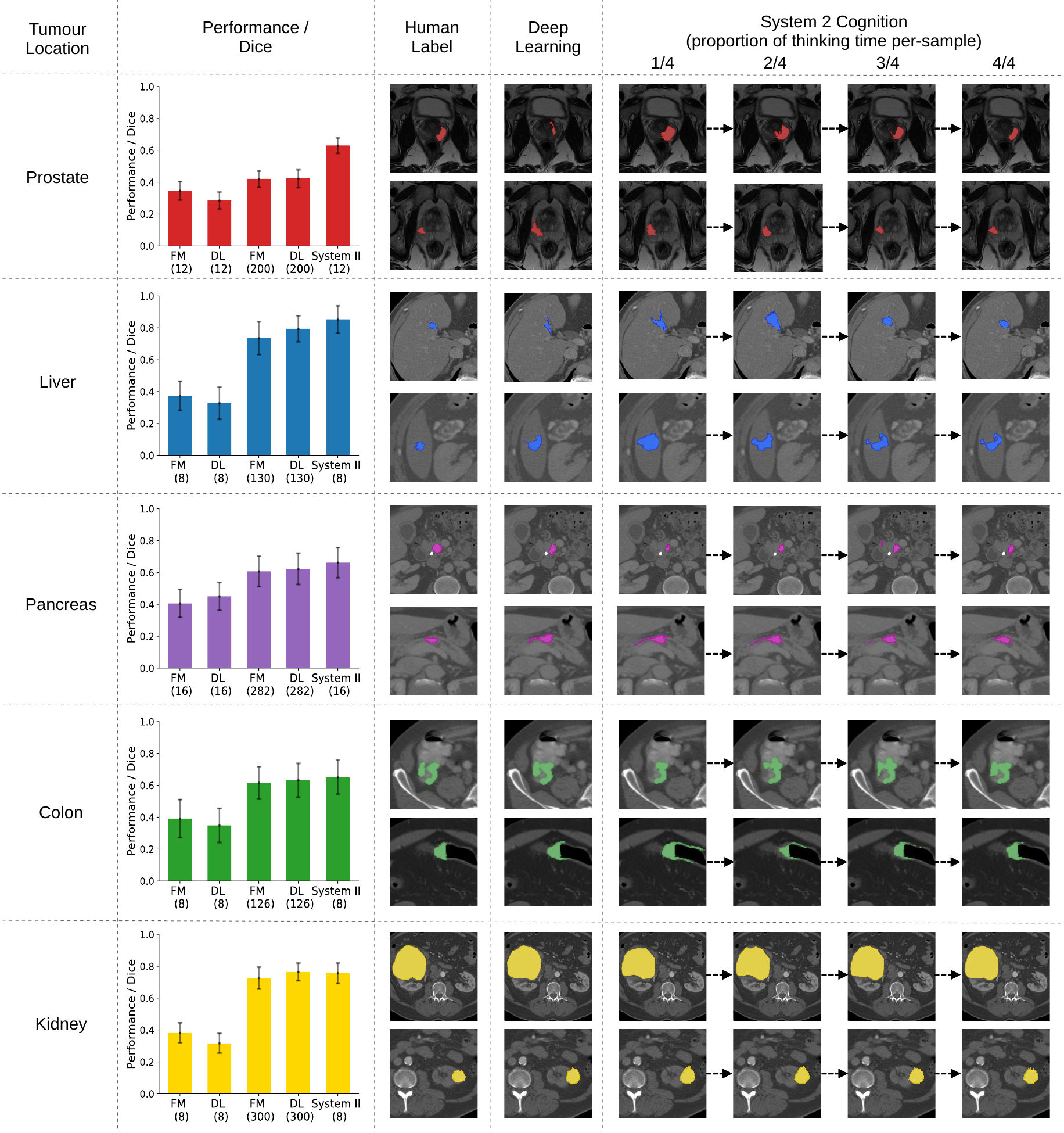}
    \caption{Performance of System II thinking compared with deep learning (DL) and foundation models (FM) in five cancer localisation tasks for different organs and image types. The left column shows statistically significant increases in cancer localisation performance compared to DL and FM with equivalent training data (indicated by the numbers in brackets below), and better or at least comparable performance even when DL and FM use substantially more training data. The remaining columns provide qualitative examples demonstrating the improvement as the inference-time compute increases, further discussed in the text.}
    \label{fig:performance_qual}
\end{figure}

\FloatBarrier

System II sets state-of-the-art performance for cancer localisation tasks: Fig. \ref{fig:performance_qual} shows that System II thinking achieved new state-of-the-art performance for cancer segmentation across all five distinct organs. 
The System I module was trained on abdomino-pelvic CT and MR images, from 955 3D human-labelled images, for over 13 anatomical structures, similar to prior work \cite{saeed2024active} (summarised in Table 1 in supplementary materials). This is also markedly less than the pre-training data sets required in recent image foundation models \cite{ma2024segment}. This was then adapted using few labelled samples (approximately 8-16, specified where appropriate) to the unseen target tasks of cancer localisation on medical images. The System II thinking module was evaluated for each of the five clinical tasks of localising cancer on MR or CT images from real cancer patients. 
Compared to the previous best methods, the System II approach outperformed (percentage points reported here) cancer segmentation in the prostate \cite{saeed2022image, saeed2022image_media, yan2022impact, pocius2024weakly} by 20.7\%, in the liver \cite{liu2023clip} by 5.9\%, in the pancreas \cite{liu2023clip} by 3.9\% and in the colon \cite{liu2023clip} by 2.0\%, with all improvements being statistically significant (all $p-values<$0.001, paired Student's t-test at a significance level of $\alpha$=0.05). It is considered equivalent to the best-performing method for kidney cancer segmentation \cite{myronenko2023automated}. For this task, statistical significance was not found for the comparison ($p-value=$0.051), with the intra-observer \cite{heller2019kits19} and inter-observer \cite{veiga2022comparative} variabilities reported to be high, being similar to that between human and machine segmentation \cite{veiga2022comparative, heller2019kits19}, which limit the highest measurable performance for any manual or automated segmentation method, due to the limited detectable effect size.

System II thinking significantly outperforms foundation models: Fig. \ref{fig:performance_qual} also shows substantial improvements using System II thinking with only 8–16 labelled samples, compared to foundation models \cite{oquab2023dinov2, butoi2023universeg} fine-tuned with the same or more (100-500) labelled samples. System II, using only 8-16 labelled samples, improves tumour segmentation over foundation models, using hundreds of labelled samples, by 20.9\% in the prostate, 11.8\% in the liver, 5.5\% in the pancreas, 3.6\% in the colon and 3.1\% in the kidney, with all improvements being statistically significant (all $p-values<$0.001).

Numerical results for all comparisons, for cancer localisation, are presented in Table 3 in the supplementary materials.

System II thinking outperforms radiologists with respect to independent histopathology ground-truth: Cancer presence in a subject can be identified through medical images and verified through an independent histopathology examination, for example, via a biopsy procedure. The classification of whether a patient has cancer or not, based on the medical image, has potential to allow non-invasive diagnoses in lieu of any invasive histology-based confirmations. Thus, for the task of classifying cancer presence in patients using only medical images, the System II approach was compared with double-blind radiologists performing the same task. The patient-wise cancer presence classifications were verified, in terms of sensitivity and specificity, against histopathology examination results. The procedure for this evaluation is outlined in `Retrospective evaluation against human experts for prostate cancer detection' in the supplementary materials.
For this patient-wise classification, System II offered improved sensitivity (with 95\% confidence interval) of 0.911 (0.880, 0.934) and specificity of 0.655 (0.576, 0.727) compared to the expert radiologist performance~\cite{ahmed2017diagnostic} with a sensitivity of 0.88 (0.84–0.91) and a specificity of 0.45 (0.39–0.51), for the same task on the same data set.

\begin{figure}[!ht]
    \centering
    \includegraphics[width=0.98\linewidth]{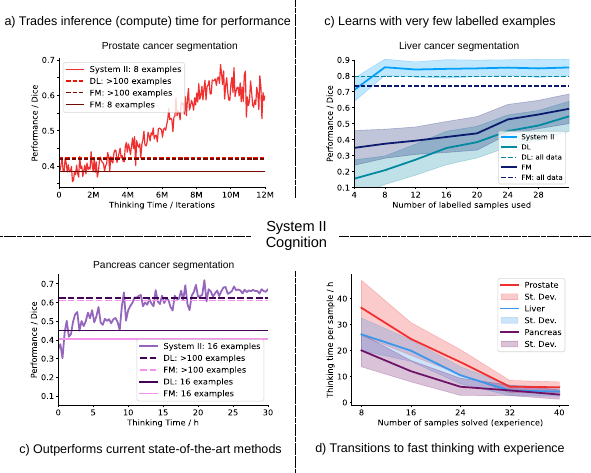}
    \caption{System II thinking: a) improves performance with increasing thinking time, compared to supervised deep learning (DL) and foundation models (FM) with fixed inference time (measured in the unit of millions of optimisation iterations); b) outperforms the current best-performing deep learning (DL) methods, using substantially fewer labelled task-specific samples; c) attains peak performance with 8 labelled samples, compared to other methods requiring more than one hundred labelled samples; d) transitions tasks to fast thinking, reducing thinking time as more unlabelled samples are fully refined (or solved) by System II thinking. Performance in plotted lines is the average across 20 samples from the test set. Complete results for all cancer types can be found in the `Comparison to existing methods for cancer localisation' section in the supplementary materials.}
    \label{fig:sys_2_features}
\end{figure}

\FloatBarrier

\subsubsection*{The abilities of the proposed inference-time compute framework}

Our experiments focus on representative data-scarce vision tasks that are clinically and scientifically important, where conventional deep learning often struggles. We do not aim to cover the full breadth of visual reasoning. Instead, we show that inference-time computation, via competitive hypothesis refinement, can reliably improve performance when additional labelled data are limited or unavailable. These results illustrate a broader principle: increasing computation at inference can serve as an effective alternative to increasing training data, with potential to extend to other visual tasks in future work.

Trading inference-time compute for machine performance:
Fig. \ref{fig:sys_2_features}a-b) demonstrate a key feature of System II cognition - the ability to improve performance by utilising inference-time compute, to a much greater extent than alternative learning systems. This is shown as performance improves with a substantially longer period of increased thinking time, when we allow the System II module to auto-compete in an iterative manner, refining the solutions as the competition progresses. In this iterative refinement process, the space of performance improvement becomes much more expandable, compared with that from other single-pass or few-shot learning methods such as foundation models that have a distinct aim of being data-efficient with fixed inference time. 

Outperforming fixed-thinking-time System I approaches:
Fig. \ref{fig:sys_2_features}a-b) show that the System II approach outperforms the currently best approaches in computer-vision (full details and justifications are outlined in Table 3 in the supplementary materials), which all follow a fixed-thinking-time System I paradigm. The performance improvement can be seen despite the System II approach using less than ten times the amount of data required by the solitary System I data-driven approaches.

Learning with very few labelled samples:
Fig. \ref{fig:sys_2_features}c) shows that System II thinking achieves its peak performance using only 8 labelled samples. This trend, of achieving peak performance even with a small number of labelled samples, was observed for all tested applications, for all cancer localisation tasks, with only 8-16 samples (as shown in Table 3 in the supplementary materials). We report the System II performances at their peaks and compare these, not only to other methods (deep learning and foundation models) using equivalent amounts of data, but also to their converged peak performances which require hundreds of samples.

Transitioning from slow System II to fast System I cognition:
Fig. \ref{fig:sys_2_features}d) shows the ability for the System II module to transition tasks to fast System I cognition. The System II module requires thinking time to refine solutions until no improvements are possible in the distribution-discriminator score, at which point a solution is considered fully-refined and thus solved. This solved solution is used to re-train the System I module, which enables better initial solutions, and more accurate distribution-discriminator scores, for new samples and reduces subsequent thinking time in the System II stage. The ability of reducing thinking time while maintaining the same performance, without any new human-labelled data is demonstrated in Fig. \ref{fig:sys_2_features}d), where the inference time in hours decreases as more samples are fully refined by the System II module. Note that we did not observe any cases where System II is no longer required (sufficient experience to reduce System II thinking time to zero), possibly due to the complexity and difficulty of the cancer localisation task.

Exploring explicit reasoning in machine System II cognition:
The intermediate solutions from the refinement process in System II thinking, presented in Fig. \ref{fig:performance_qual}, show patterns of explicit reasoning in our algorithm that mimic the ability of human System II cognition. All refinements in the competition that led to higher distribution-discriminator scores, compared to the opposing lower-scoring agent, can be visualised to draw sample-specific insights and determine where and how the refinements improved solutions. As a specific example, the first row in Fig. \ref{fig:performance_qual} shows that the prostate gland cancerous lesion had lower intensity values and a conspicuous pattern in the peripheral zone distinct from the surrounding healthy tissue. This was challenging to identify due to close proximity of the lesion to the transitional zone which also has lower intensity compared to the peripheral zone. This may likely be the reason that the deep learning systems, without System II capabilities, deemed a significant portion of the transitional zone to be a false positive pathological region. Similar gradual local adjustments are seen in other examples, e.g., in the first sample for liver tumour segmentation where the tumour is correctly identified in System II, despite its close proximity to the falciform ligament with a similar intensity. The correct identification of only the tumour region in System II thinking, guided by the distribution scores, along with the trajectory of refinements, demonstrates its ability to overcome such challenging ambiguities in separating anatomical and pathological voxels, using inference-time refinement.

\section*{Discussion}

The proposed dual-process framework enables performance in machine vision tasks to improve with increasing thinking time, through an iterative refinement of solutions where multiple competing solution strategies are considered before enacting the refinements - much like human thought processes. This contrasts with conventional deep learning methods, where post-training gains typically require task-specific annotations and the thinking/ inference time is fixed. In vision tasks, even recently developed foundation models and meta-learning approaches, designed for efficient data use, cannot improve through extended inference time. By contrast, System II demonstrates significant performance gains in real-world vision tasks such as cancer detection on medical images, purely by increasing thinking time. 
This highlights the potential to automate many challenging tasks through inference-time compute strategies, particularly when large amounts of labelled data are expensive or impossible to obtain, e.g., in data-scare domains such as rare disease identification and galaxy classification in astronomy. With compute power becoming more readily accessible, such approaches have potential to improve a variety of challenging real-world tasks, potentially even when large data-sets are available.

We emphasise that the use of dual-process terminology serves only as a high-level conceptual analogy for separating fast feed-forward prediction from slower unbounded iterative refinement, and is not intended as an emulated cognitive model. Our empirical work shows the performance improvements, regardless of the availability of task-specific data, possible by enabling inference-time compute in vision tasks where data are scarce.

As noted above, our experiments focus on a few vision tasks, chosen because they represent clinically and scientifically important settings where labelled data are scarce and conventional deep-learning methods underperform. We do not claim to address the full breadth of visual reasoning, which spans diverse domains and task families. Instead, our results demonstrate that inference-time compute or reasoning, implemented through competitive hypothesis refinement, can reliably improve performance in settings where additional labelled data are difficult or impossible to obtain. We view these studies as exemplars of a broader principle, where extending computation at inference can serve as a powerful alternative to expanding training datasets, and may generalise to a wider range of visual tasks in future work.

Although the proposed framework demonstrated strong performance across the evaluated tasks, several potential failure modes are worth noting. First, because System II optimisation uses the discriminator score as a learned proxy for the true task objective, occasional discrepancies between the discriminator and the true metric (e.g., Dice or IoU) could lead to refinements that improve the proxy score without improving the underlying segmentation quality. Second, when the initial System I prediction is extremely poor, a larger number of refinement iterations may be required to approach an accurate solution, potentially exceeding practical inference-time compute budgets in our applications. Third, the discriminator is adapted using a small set of labelled samples from the target domain, and if this adaptation set is unrepresentative (for example containing predominantly large lesions), performance on substantially different structures may be affected. Cross-task pre-training substantially mitigates this risk by making the reasoning process less sensitive to the particular small adaptation set used. While some performance degradation may still arise in the presence of highly atypical outlier data, this is unlikely to affect a whole adaptation set in practice, and can be further minimised through appropriate adaptation-set design and selection.

The proposed refinement process shares some similarities with iterative denoising in diffusion models, however, the underlying mechanism is fundamentally different. Diffusion models learn a fixed reverse process from large datasets that maps noisy inputs to the data distribution, and apply this process at inference. In contrast, our framework performs per-sample optimisation guided by a learned discriminator, where candidate refinements are iteratively generated, evaluated, and selected. This enables adaptive inference-time refinement conditioned on the specific sample and task objective, allowing direct per-sample optimisation rather than following a predefined denoising path.

While we model System I thinking as an adaptable neural network trained across various tasks, there is potential to consider existing foundation models in this framework and improve their predictions via the proposed inference-time refinement through System II thinking. It is also noteworthy that while our competitive refinement framework for enabling System II thinking provides a general mechanism applicable to various vision problems, some tasks may benefit from different search-based formulations or guiding inference-time compute through human feedback formulated as a reward signal.

In addition to exploring a new research direction in modelling anthropomorphic reasoning and decision-making in machines, the System II cognitive process opens a new data-efficient, performance-improving mechanism, applicable in common real-world scenarios where acquiring labelled data to improve performance may be expensive or prohibitive.

\section*{Methods}

\begin{figure}[!ht]
    \centering
    \includegraphics[width=0.98\linewidth]{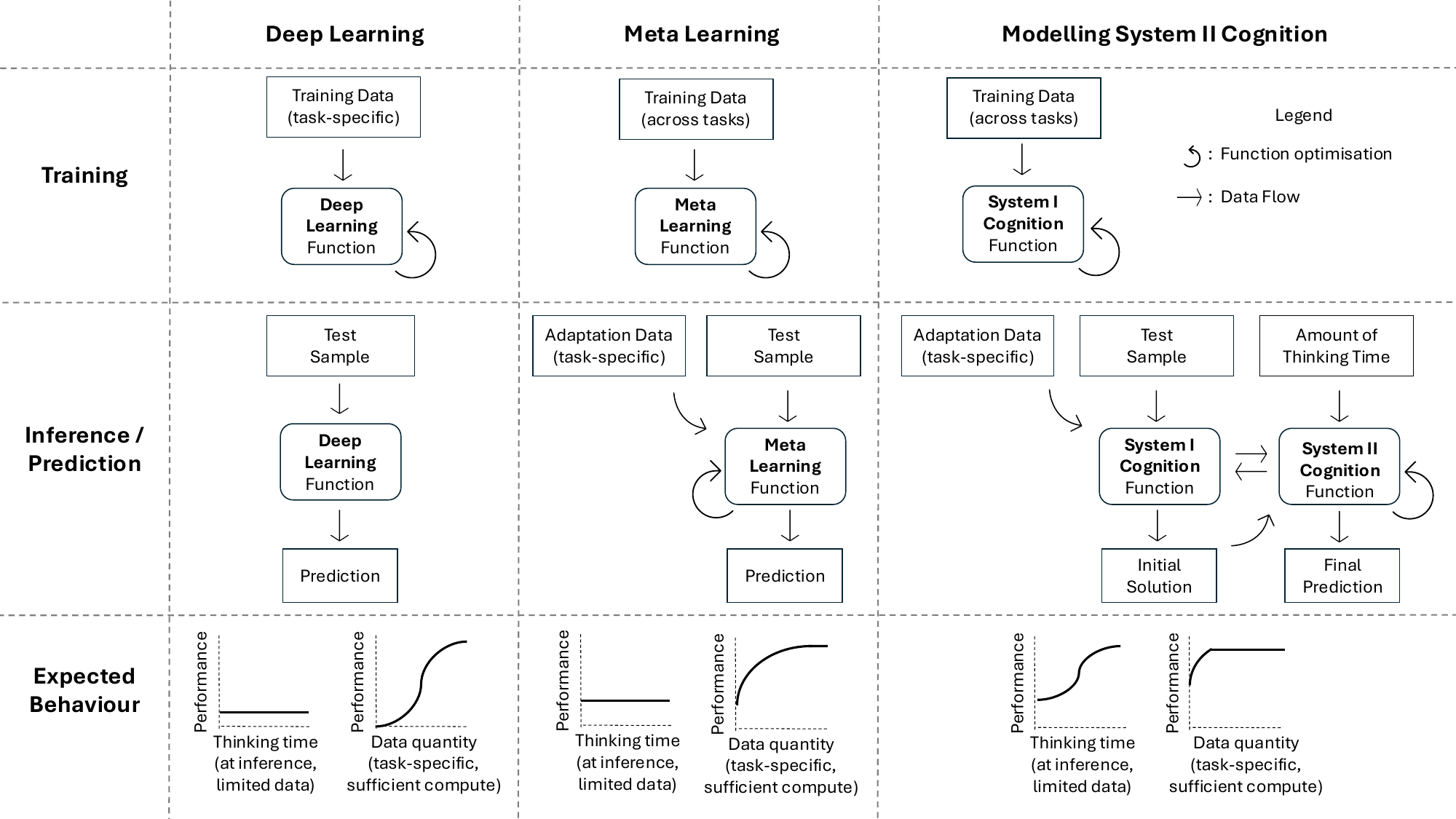}
    \caption{Different learning strategies are compared: deep learning for a single forward pass prediction, meta-learning for adapting to new tasks (akin to fine-tuning pre-trained foundation models), and the proposed System II cognition for allowing inference-time compute. Details of these are described in text.}
    \label{fig:learning_comp}
\end{figure}

An overview of the proposed approach, along with comparison to conventional approaches, is presented in Fig. \ref{fig:learning_comp}.

For our method, we propose a System I module that consists of two neural networks trained adversarially \cite{goodfellow2020generative} across a variety of tasks, using human-labelled data. This mimics the experience-driven System I thinking in human psychology. The first network in this module is a task-predictor, which produces a solution (such as localising an object) given an input image. The second network is a distribution-discriminator which measures the likelihood of solutions being correct, by leveraging pre-learned knowledge from a variety of human-labelled data across multiple tasks. In the adversarial training, human labels are regarded as correct solutions and task-predictor solutions are regarded as incorrect. The aim during training is for the task-predictor to produce solutions close to human labels, and the distribution-discriminator to differentiate between human and task-predictor solutions. At inference for a new task, the task-predictor can be adapted to produce an initial solution for inference-time reasoning in the System II module, while the distribution-discriminator can be adapted to reward refinement of solutions towards the direction of correct solutions. This adaptation uses a small number of human-labelled samples (8-16 in our experiments) to adapt both the task-predictor and distribution-discriminator prior to System II refinement to ensure task-specific predictions for the new task. The adapted distribution-discriminator in our framework is used to guide System II refinements, as it is considered more easily generalisable than the task-predictor (with higher-dimensional outputs such as pixel-level segmentation). This offers an interesting explanation for the potential performance improvement, consistent with findings from both psychology \cite{yun2013exploring} (e.g. it is easy to recognise good writing vs bad but it is difficult to produce a well-written piece-of-work) and machine learning literature \cite{nichol2018first} (analogous to the discriminative-vs-generative comparison, also due to the size of solution space), as well as our own experiments (see the `Ablation studies' section in supplementary materials).

The System II module aims to refine the solution for a specific sample from the target task. 
We implement the refinement as a flexible self-play reinforcement learning \cite{silver2018general, saeed2024competing} process, where two competing neural network agents propose refinements to an initial solution produced by the task-predictor. The distribution-discriminator scores the competing refinements by measuring the likelihood of correctness for each. The refinement with a higher score is rewarded and enacted. The competing networks are then updated, using this distribution-discriminator score-based reward, to produce progressively better competing refinements in subsequent iterations. The process repeats and the solution is iteratively refined until no further improvement is possible, as measured by the distribution-discriminator score. This iterative refinement-proposal and evaluation facilitates a dynamic improvement process where each refined solution becomes the starting point for further optimisation. This iterative self-play process thus facilitates the System II thinking module to improve solutions with added thinking time.

The final refined solution from the System II module can be used to re-train the System I module and transition tasks to fast thinking as experience grows. So, as more samples are exposed to System II thinking, the new task-specific experience of the System I module grows, which enables better initial solutions with increasing experience. This eventually reduces the time and reliance on the auto-competing System II stage. However, whether or not a task can be fully transitioned to System I, requiring no refinement in the System II stage, depends on the complexity of the task and the ability to cover task variability sufficiently. As demonstrated in the results, certain tasks may have an upper-bound of performance, which can be achieved either through extended reasoning in System II thinking or through an expansive dataset covering the task variability. Nevertheless, complex tasks may always require System II thinking to obtain optimal performance, given finite training data unable to cover task variability.

The System II framework also models the explicit reasoning in System II thinking. Due to the iterative nature of the solution-refining System II module, the full trajectory of refinement provides a record of where and how solutions were improved, enhancing interpretability and offering insights into justifications for specific final solutions.

The framework is summarised in Fig. \ref{fig:method_overview}a) and \ref{fig:method_overview}b). Detailed methodological descriptions for implementing and training of the functions, in both modules, are presented in the `Methodological description' section of the supplementary materials and open-sourced at \url{github.com/s-sd/system-II-vision} \cite{shaheer_u_saeed_2026_19331690}.

\newpage

\section*{Data Availability}
The MNIST data used in this study were created within \cite{shaheer_u_saeed_2026_19331690} available at \url{github.com/s-sd/system-II-vision}. The ImageNet S data is available openly via \cite{gao2022large}, access can be obtained by following \url{github.com/LUSSeg/ImageNet-S}. The medical imaging data is all openly available with details provided in the Supplementary Table S3; and an example data loader and minimal set provided in \cite{shaheer_u_saeed_2026_19331690} available at \url{github.com/s-sd/system-II-vision}.

\section*{Code Availability}
The code is available openly via  \cite{shaheer_u_saeed_2026_19331690}, available at \url{github.com/s-sd/system-II-vision}.

\newpage

\bibliographystyle{alpha}

\begin{thebibliography}{78}
\ifx \bisbn   \undefined \def \bisbn  #1{ISBN #1}\fi
\ifx \binits  \undefined \def \binits#1{#1}\fi
\ifx \bauthor  \undefined \def \bauthor#1{#1}\fi
\ifx \batitle  \undefined \def \batitle#1{#1}\fi
\ifx \bjtitle  \undefined \def \bjtitle#1{#1}\fi
\ifx \bvolume  \undefined \def \bvolume#1{\textbf{#1}}\fi
\ifx \byear  \undefined \def \byear#1{#1}\fi
\ifx \bissue  \undefined \def \bissue#1{#1}\fi
\ifx \bfpage  \undefined \def \bfpage#1{#1}\fi
\ifx \blpage  \undefined \def \blpage #1{#1}\fi
\ifx \burl  \undefined \def \burl#1{\textsf{#1}}\fi
\ifx \doiurl  \undefined \def \doiurl#1{\url{https://doi.org/#1}}\fi
\ifx \betal  \undefined \def \betal{\textit{et al.}}\fi
\ifx \binstitute  \undefined \def \binstitute#1{#1}\fi
\ifx \binstitutionaled  \undefined \def \binstitutionaled#1{#1}\fi
\ifx \bctitle  \undefined \def \bctitle#1{#1}\fi
\ifx \beditor  \undefined \def \beditor#1{#1}\fi
\ifx \bpublisher  \undefined \def \bpublisher#1{#1}\fi
\ifx \bbtitle  \undefined \def \bbtitle#1{#1}\fi
\ifx \bedition  \undefined \def \bedition#1{#1}\fi
\ifx \bseriesno  \undefined \def \bseriesno#1{#1}\fi
\ifx \blocation  \undefined \def \blocation#1{#1}\fi
\ifx \bsertitle  \undefined \def \bsertitle#1{#1}\fi
\ifx \bsnm \undefined \def \bsnm#1{#1}\fi
\ifx \bsuffix \undefined \def \bsuffix#1{#1}\fi
\ifx \bparticle \undefined \def \bparticle#1{#1}\fi
\ifx \barticle \undefined \def \barticle#1{#1}\fi

\ifx \bconfdate \undefined \def \bconfdate #1{#1}\fi
\ifx \botherref \undefined \def \botherref #1{#1}\fi
\ifx \bchapter \undefined \def \bchapter#1{#1}\fi
\ifx \bbook \undefined \def \bbook#1{#1}\fi
\ifx \bcomment \undefined \def \bcomment#1{#1}\fi
\ifx \oauthor \undefined \def \oauthor#1{#1}\fi
\ifx \citeauthoryear \undefined \def \citeauthoryear#1{#1}\fi
\ifx \endbibitem  \undefined \def \endbibitem {}\fi
\ifx \bconflocation  \undefined \def \bconflocation#1{#1}\fi
\ifx \arxivurl  \undefined \def \arxivurl#1{\textsf{#1}}\fi
\csname PreBibitemsHook\endcsname

\bibitem[\protect\citeauthoryear{Kahneman}{2003}]{kahneman2003maps}
\begin{barticle}
\bauthor{\bsnm{Kahneman}, \binits{D.}}:
\batitle{Maps of bounded rationality: Psychology for behavioral economics}.
\bjtitle{American economic review}
\bvolume{93}(\bissue{5}),
\bfpage{1449}--\blpage{1475}
(\byear{2003})
\end{barticle}
\endbibitem

\bibitem[\protect\citeauthoryear{Evans}{1984}]{evans1984heuristic}
\begin{barticle}
\bauthor{\bsnm{Evans}, \binits{J.S.B.}}:
\batitle{Heuristic and analytic processes in reasoning}.
\bjtitle{British Journal of Psychology}
\bvolume{75}(\bissue{4}),
\bfpage{451}--\blpage{468}
(\byear{1984})
\end{barticle}
\endbibitem

\bibitem[\protect\citeauthoryear{Kahneman}{2012}]{kahneman20122}
\begin{botherref}
\oauthor{\bsnm{Kahneman}, \binits{D.}}:
Of 2 minds: How fast and slow thinking shape perception and choice [excerpt].
Scientific American
\textbf{15}
(2012)
\end{botherref}
\endbibitem

\bibitem[\protect\citeauthoryear{Anthony et~al.}{2017}]{anthony2017thinking}
\begin{botherref}
\oauthor{\bsnm{Anthony}, \binits{T.}},
\oauthor{\bsnm{Tian}, \binits{Z.}},
\oauthor{\bsnm{Barber}, \binits{D.}}:
Thinking fast and slow with deep learning and tree search.
Advances in neural information processing systems
\textbf{30}
(2017)
\end{botherref}
\endbibitem

\bibitem[\protect\citeauthoryear{Guo et~al.}{2025}]{deepseek_r1}
\begin{botherref}
\oauthor{\bsnm{Guo}, \binits{D.}},
\oauthor{\bsnm{Yang}, \binits{D.}},
\oauthor{\bsnm{Zhang}, \binits{H.}},
\oauthor{\bsnm{Song}, \binits{J.}},
\oauthor{\bsnm{Zhang}, \binits{R.}},
\oauthor{\bsnm{Xu}, \binits{R.}},
\oauthor{\bsnm{Zhu}, \binits{Q.}},
\oauthor{\bsnm{Ma}, \binits{S.}},
\oauthor{\bsnm{Wang}, \binits{P.}},
\oauthor{\bsnm{Bi}, \binits{X.}}, et al.:
Deepseek-r1: Incentivizing reasoning capability in llms via reinforcement learning.
arXiv preprint arXiv:2501.12948
(2025)
\end{botherref}
\endbibitem

\bibitem[\protect\citeauthoryear{Wei et~al.}{2022}]{wei2022chain}
\begin{barticle}
\bauthor{\bsnm{Wei}, \binits{J.}},
\bauthor{\bsnm{Wang}, \binits{X.}},
\bauthor{\bsnm{Schuurmans}, \binits{D.}},
\bauthor{\bsnm{Bosma}, \binits{M.}},
\bauthor{\bsnm{Xia}, \binits{F.}},
\bauthor{\bsnm{Chi}, \binits{E.}},
\bauthor{\bsnm{Le}, \binits{Q.V.}},
\bauthor{\bsnm{Zhou}, \binits{D.}}, \betal:
\batitle{Chain-of-thought prompting elicits reasoning in large language models}.
\bjtitle{Advances in neural information processing systems}
\bvolume{35},
\bfpage{24824}--\blpage{24837}
(\byear{2022})
\end{barticle}
\endbibitem

\bibitem[\protect\citeauthoryear{Evans}{2003}]{evans2003two}
\begin{barticle}
\bauthor{\bsnm{Evans}, \binits{J.S.B.}}:
\batitle{In two minds: dual-process accounts of reasoning}.
\bjtitle{Trends in cognitive sciences}
\bvolume{7}(\bissue{10}),
\bfpage{454}--\blpage{459}
(\byear{2003})
\end{barticle}
\endbibitem

\bibitem[\protect\citeauthoryear{Litjens et~al.}{2017}]{litjens2017survey}
\begin{barticle}
\bauthor{\bsnm{Litjens}, \binits{G.}},
\bauthor{\bsnm{Kooi}, \binits{T.}},
\bauthor{\bsnm{Bejnordi}, \binits{B.E.}},
\bauthor{\bsnm{Setio}, \binits{A.A.A.}},
\bauthor{\bsnm{Ciompi}, \binits{F.}},
\bauthor{\bsnm{Ghafoorian}, \binits{M.}},
\bauthor{\bsnm{Van Der~Laak}, \binits{J.A.}},
\bauthor{\bsnm{Van~Ginneken}, \binits{B.}},
\bauthor{\bsnm{S{\'a}nchez}, \binits{C.I.}}:
\batitle{A survey on deep learning in medical image analysis}.
\bjtitle{Medical image analysis}
\bvolume{42},
\bfpage{60}--\blpage{88}
(\byear{2017})
\end{barticle}
\endbibitem

\bibitem[\protect\citeauthoryear{Bhattacharya et~al.}{2022}]{bhattacharya2022review}
\begin{barticle}
\bauthor{\bsnm{Bhattacharya}, \binits{I.}},
\bauthor{\bsnm{Khandwala}, \binits{Y.S.}},
\bauthor{\bsnm{Vesal}, \binits{S.}},
\bauthor{\bsnm{Shao}, \binits{W.}},
\bauthor{\bsnm{Yang}, \binits{Q.}},
\bauthor{\bsnm{Soerensen}, \binits{S.J.}},
\bauthor{\bsnm{Fan}, \binits{R.E.}},
\bauthor{\bsnm{Ghanouni}, \binits{P.}},
\bauthor{\bsnm{Kunder}, \binits{C.A.}},
\bauthor{\bsnm{Brooks}, \binits{J.D.}}, \betal:
\batitle{A review of artificial intelligence in prostate cancer detection on imaging}.
\bjtitle{Therapeutic advances in urology}
\bvolume{14},
\bfpage{17562872221128791}
(\byear{2022})
\end{barticle}
\endbibitem

\bibitem[\protect\citeauthoryear{Koh et~al.}{2022}]{koh2022artificial}
\begin{barticle}
\bauthor{\bsnm{Koh}, \binits{D.-M.}},
\bauthor{\bsnm{Papanikolaou}, \binits{N.}},
\bauthor{\bsnm{Bick}, \binits{U.}},
\bauthor{\bsnm{Illing}, \binits{R.}},
\bauthor{\bsnm{Kahn~Jr}, \binits{C.E.}},
\bauthor{\bsnm{Kalpathi-Cramer}, \binits{J.}},
\bauthor{\bsnm{Matos}, \binits{C.}},
\bauthor{\bsnm{Mart{\'\i}-Bonmat{\'\i}}, \binits{L.}},
\bauthor{\bsnm{Miles}, \binits{A.}},
\bauthor{\bsnm{Mun}, \binits{S.K.}}, \betal:
\batitle{Artificial intelligence and machine learning in cancer imaging}.
\bjtitle{Communications Medicine}
\bvolume{2}(\bissue{1}),
\bfpage{133}
(\byear{2022})
\end{barticle}
\endbibitem

\bibitem[\protect\citeauthoryear{S{\"o}rqvist and Marsh}{2015}]{sorqvist2015concentration}
\begin{barticle}
\bauthor{\bsnm{S{\"o}rqvist}, \binits{P.}},
\bauthor{\bsnm{Marsh}, \binits{J.E.}}:
\batitle{How concentration shields against distraction}.
\bjtitle{Current directions in psychological science}
\bvolume{24}(\bissue{4}),
\bfpage{267}--\blpage{272}
(\byear{2015})
\end{barticle}
\endbibitem

\bibitem[\protect\citeauthoryear{Vabalas et~al.}{2019}]{vabalas2019machine}
\begin{barticle}
\bauthor{\bsnm{Vabalas}, \binits{A.}},
\bauthor{\bsnm{Gowen}, \binits{E.}},
\bauthor{\bsnm{Poliakoff}, \binits{E.}},
\bauthor{\bsnm{Casson}, \binits{A.J.}}:
\batitle{Machine learning algorithm validation with a limited sample size}.
\bjtitle{PloS one}
\bvolume{14}(\bissue{11}),
\bfpage{0224365}
(\byear{2019})
\end{barticle}
\endbibitem

\bibitem[\protect\citeauthoryear{Wang et~al.}{2020}]{wang2020generalizing}
\begin{barticle}
\bauthor{\bsnm{Wang}, \binits{Y.}},
\bauthor{\bsnm{Yao}, \binits{Q.}},
\bauthor{\bsnm{Kwok}, \binits{J.T.}},
\bauthor{\bsnm{Ni}, \binits{L.M.}}:
\batitle{Generalizing from a few examples: A survey on few-shot learning}.
\bjtitle{ACM computing surveys (csur)}
\bvolume{53}(\bissue{3}),
\bfpage{1}--\blpage{34}
(\byear{2020})
\end{barticle}
\endbibitem

\bibitem[\protect\citeauthoryear{Ravi and Larochelle}{2016}]{ravi2016optimization}
\begin{bchapter}
\bauthor{\bsnm{Ravi}, \binits{S.}},
\bauthor{\bsnm{Larochelle}, \binits{H.}}:
\bctitle{Optimization as a model for few-shot learning}.
In: \bbtitle{International Conference on Learning Representations}
(\byear{2016})
\end{bchapter}
\endbibitem

\bibitem[\protect\citeauthoryear{Snell et~al.}{2017}]{snell2017prototypical}
\begin{botherref}
\oauthor{\bsnm{Snell}, \binits{J.}},
\oauthor{\bsnm{Swersky}, \binits{K.}},
\oauthor{\bsnm{Zemel}, \binits{R.}}:
Prototypical networks for few-shot learning.
Advances in neural information processing systems
\textbf{30}
(2017)
\end{botherref}
\endbibitem

\bibitem[\protect\citeauthoryear{Kirillov et~al.}{2023}]{kirillov2023segment}
\begin{bchapter}
\bauthor{\bsnm{Kirillov}, \binits{A.}},
\bauthor{\bsnm{Mintun}, \binits{E.}},
\bauthor{\bsnm{Ravi}, \binits{N.}},
\bauthor{\bsnm{Mao}, \binits{H.}},
\bauthor{\bsnm{Rolland}, \binits{C.}},
\bauthor{\bsnm{Gustafson}, \binits{L.}},
\bauthor{\bsnm{Xiao}, \binits{T.}},
\bauthor{\bsnm{Whitehead}, \binits{S.}},
\bauthor{\bsnm{Berg}, \binits{A.C.}},
\bauthor{\bsnm{Lo}, \binits{W.-Y.}}, \betal:
\bctitle{Segment anything}.
In: \bbtitle{Proceedings of the IEEE/CVF International Conference on Computer Vision},
pp. \bfpage{4015}--\blpage{4026}
(\byear{2023})
\end{bchapter}
\endbibitem

\bibitem[\protect\citeauthoryear{Ma et~al.}{2024}]{ma2024segment}
\begin{barticle}
\bauthor{\bsnm{Ma}, \binits{J.}},
\bauthor{\bsnm{He}, \binits{Y.}},
\bauthor{\bsnm{Li}, \binits{F.}},
\bauthor{\bsnm{Han}, \binits{L.}},
\bauthor{\bsnm{You}, \binits{C.}},
\bauthor{\bsnm{Wang}, \binits{B.}}:
\batitle{Segment anything in medical images}.
\bjtitle{Nature Communications}
\bvolume{15}(\bissue{1}),
\bfpage{654}
(\byear{2024})
\end{barticle}
\endbibitem

\bibitem[\protect\citeauthoryear{Zhou et~al.}{2023}]{zhou2023foundation}
\begin{barticle}
\bauthor{\bsnm{Zhou}, \binits{Y.}},
\bauthor{\bsnm{Chia}, \binits{M.A.}},
\bauthor{\bsnm{Wagner}, \binits{S.K.}},
\bauthor{\bsnm{Ayhan}, \binits{M.S.}},
\bauthor{\bsnm{Williamson}, \binits{D.J.}},
\bauthor{\bsnm{Struyven}, \binits{R.R.}},
\bauthor{\bsnm{Liu}, \binits{T.}},
\bauthor{\bsnm{Xu}, \binits{M.}},
\bauthor{\bsnm{Lozano}, \binits{M.G.}},
\bauthor{\bsnm{Woodward-Court}, \binits{P.}}, \betal:
\batitle{A foundation model for generalizable disease detection from retinal images}.
\bjtitle{Nature}
\bvolume{622}(\bissue{7981}),
\bfpage{156}--\blpage{163}
(\byear{2023})
\end{barticle}
\endbibitem

\bibitem[\protect\citeauthoryear{Moor et~al.}{2023}]{moor2023foundation}
\begin{barticle}
\bauthor{\bsnm{Moor}, \binits{M.}},
\bauthor{\bsnm{Banerjee}, \binits{O.}},
\bauthor{\bsnm{Abad}, \binits{Z.S.H.}},
\bauthor{\bsnm{Krumholz}, \binits{H.M.}},
\bauthor{\bsnm{Leskovec}, \binits{J.}},
\bauthor{\bsnm{Topol}, \binits{E.J.}},
\bauthor{\bsnm{Rajpurkar}, \binits{P.}}:
\batitle{Foundation models for generalist medical artificial intelligence}.
\bjtitle{Nature}
\bvolume{616}(\bissue{7956}),
\bfpage{259}--\blpage{265}
(\byear{2023})
\end{barticle}
\endbibitem

\bibitem[\protect\citeauthoryear{Mazurowski et~al.}{2023}]{mazurowski2023segment}
\begin{barticle}
\bauthor{\bsnm{Mazurowski}, \binits{M.A.}},
\bauthor{\bsnm{Dong}, \binits{H.}},
\bauthor{\bsnm{Gu}, \binits{H.}},
\bauthor{\bsnm{Yang}, \binits{J.}},
\bauthor{\bsnm{Konz}, \binits{N.}},
\bauthor{\bsnm{Zhang}, \binits{Y.}}:
\batitle{Segment anything model for medical image analysis: an experimental study}.
\bjtitle{Medical Image Analysis}
\bvolume{89},
\bfpage{102918}
(\byear{2023})
\end{barticle}
\endbibitem

\bibitem[\protect\citeauthoryear{Xie et~al.}{2024}]{xie2024sam}
\begin{bchapter}
\bauthor{\bsnm{Xie}, \binits{W.}},
\bauthor{\bsnm{Willems}, \binits{N.}},
\bauthor{\bsnm{Patil}, \binits{S.}},
\bauthor{\bsnm{Li}, \binits{Y.}},
\bauthor{\bsnm{Kumar}, \binits{M.}}:
\bctitle{Sam fewshot finetuning for anatomical segmentation in medical images}.
In: \bbtitle{Proceedings of the IEEE/CVF Winter Conference on Applications of Computer Vision},
pp. \bfpage{3253}--\blpage{3261}
(\byear{2024})
\end{bchapter}
\endbibitem

\bibitem[\protect\citeauthoryear{Li and Rajpurkar}{2024}]{li2024adapting}
\begin{bchapter}
\bauthor{\bsnm{Li}, \binits{K.}},
\bauthor{\bsnm{Rajpurkar}, \binits{P.}}:
\bctitle{Adapting segment anything models to medical imaging via fine-tuning without domain pretraining}.
In: \bbtitle{AAAI 2024 Spring Symposium on Clinical Foundation Models}
(\byear{2024})
\end{bchapter}
\endbibitem

\bibitem[\protect\citeauthoryear{Hosseini et~al.}{2024}]{hosseini2024v}
\begin{botherref}
\oauthor{\bsnm{Hosseini}, \binits{A.}},
\oauthor{\bsnm{Yuan}, \binits{X.}},
\oauthor{\bsnm{Malkin}, \binits{N.}},
\oauthor{\bsnm{Courville}, \binits{A.}},
\oauthor{\bsnm{Sordoni}, \binits{A.}},
\oauthor{\bsnm{Agarwal}, \binits{R.}}:
V-star: Training verifiers for self-taught reasoners.
arXiv preprint arXiv:2402.06457
(2024)
\end{botherref}
\endbibitem

\bibitem[\protect\citeauthoryear{Liu et~al.}{2025}]{liu2025can}
\begin{botherref}
\oauthor{\bsnm{Liu}, \binits{R.}},
\oauthor{\bsnm{Gao}, \binits{J.}},
\oauthor{\bsnm{Zhao}, \binits{J.}},
\oauthor{\bsnm{Zhang}, \binits{K.}},
\oauthor{\bsnm{Li}, \binits{X.}},
\oauthor{\bsnm{Qi}, \binits{B.}},
\oauthor{\bsnm{Ouyang}, \binits{W.}},
\oauthor{\bsnm{Zhou}, \binits{B.}}:
Can 1b llm surpass 405b llm? rethinking compute-optimal test-time scaling.
arXiv preprint arXiv:2502.06703
(2025)
\end{botherref}
\endbibitem

\bibitem[\protect\citeauthoryear{Su et~al.}{2024}]{su2024dualformercontrollablefastslow}
\begin{botherref}
\oauthor{\bsnm{Su}, \binits{D.}},
\oauthor{\bsnm{Sukhbaatar}, \binits{S.}},
\oauthor{\bsnm{Rabbat}, \binits{M.}},
\oauthor{\bsnm{Tian}, \binits{Y.}},
\oauthor{\bsnm{Zheng}, \binits{Q.}}:
Dualformer: Controllable Fast and Slow Thinking by Learning with Randomized Reasoning Traces
(2024).
\url{https://arxiv.org/abs/2410.09918}
\end{botherref}
\endbibitem

\bibitem[\protect\citeauthoryear{You et~al.}{2024}]{you2024biochemicalprostatecancerrecurrence}
\begin{botherref}
\oauthor{\bsnm{You}, \binits{S.}},
\oauthor{\bsnm{Adap}, \binits{S.}},
\oauthor{\bsnm{Thakur}, \binits{S.}},
\oauthor{\bsnm{Baheti}, \binits{B.}},
\oauthor{\bsnm{Bakas}, \binits{S.}}:
Biochemical Prostate Cancer Recurrence Prediction: Thinking Fast \& Slow
(2024).
\url{https://arxiv.org/abs/2409.02284}
\end{botherref}
\endbibitem

\bibitem[\protect\citeauthoryear{Ganapini et~al.}{2022}]{ganapini2022combining}
\begin{bchapter}
\bauthor{\bsnm{Ganapini}, \binits{M.B.}},
\bauthor{\bsnm{Campbell}, \binits{M.}},
\bauthor{\bsnm{Fabiano}, \binits{F.}},
\bauthor{\bsnm{Horesh}, \binits{L.}},
\bauthor{\bsnm{Lenchner}, \binits{J.}},
\bauthor{\bsnm{Loreggia}, \binits{A.}},
\bauthor{\bsnm{Mattei}, \binits{N.}},
\bauthor{\bsnm{Rossi}, \binits{F.}},
\bauthor{\bsnm{Srivastava}, \binits{B.}},
\bauthor{\bsnm{Venable}, \binits{K.B.}}, \betal:
\bctitle{Combining fast and slow thinking for human-like and efficient decisions in constrained environments.}
In: \bbtitle{NeSy},
pp. \bfpage{171}--\blpage{185}
(\byear{2022})
\end{bchapter}
\endbibitem

\bibitem[\protect\citeauthoryear{Fabiano et~al.}{2023}]{fabiano2023fast}
\begin{botherref}
\oauthor{\bsnm{Fabiano}, \binits{F.}},
\oauthor{\bsnm{Pallagani}, \binits{V.}},
\oauthor{\bsnm{Ganapini}, \binits{M.B.}},
\oauthor{\bsnm{Horesh}, \binits{L.}},
\oauthor{\bsnm{Loreggia}, \binits{A.}},
\oauthor{\bsnm{Murugesan}, \binits{K.}},
\oauthor{\bsnm{Rossi}, \binits{F.}},
\oauthor{\bsnm{Srivastava}, \binits{B.}}:
Fast and slow planning.
arXiv preprint arXiv:2303.04283
(2023)
\end{botherref}
\endbibitem

\bibitem[\protect\citeauthoryear{Pham et~al.}{2023}]{pham2023continual}
\begin{botherref}
\oauthor{\bsnm{Pham}, \binits{Q.}},
\oauthor{\bsnm{Liu}, \binits{C.}},
\oauthor{\bsnm{Hoi}, \binits{S.C.}}:
Continual learning, fast and slow.
IEEE Transactions on Pattern Analysis and Machine Intelligence
(2023)
\end{botherref}
\endbibitem

\bibitem[\protect\citeauthoryear{Qi et~al.}{2024}]{qi2024interactive}
\begin{bchapter}
\bauthor{\bsnm{Qi}, \binits{B.}},
\bauthor{\bsnm{Chen}, \binits{X.}},
\bauthor{\bsnm{Gao}, \binits{J.}},
\bauthor{\bsnm{Li}, \binits{D.}},
\bauthor{\bsnm{Liu}, \binits{J.}},
\bauthor{\bsnm{Wu}, \binits{L.}},
\bauthor{\bsnm{Zhou}, \binits{B.}}:
\bctitle{Interactive continual learning: Fast and slow thinking}.
In: \bbtitle{Proceedings of the IEEE/CVF Conference on Computer Vision and Pattern Recognition},
pp. \bfpage{12882}--\blpage{12892}
(\byear{2024})
\end{bchapter}
\endbibitem

\bibitem[\protect\citeauthoryear{Guo et~al.}{2014}]{guo2014deep}
\begin{botherref}
\oauthor{\bsnm{Guo}, \binits{X.}},
\oauthor{\bsnm{Singh}, \binits{S.}},
\oauthor{\bsnm{Lee}, \binits{H.}},
\oauthor{\bsnm{Lewis}, \binits{R.L.}},
\oauthor{\bsnm{Wang}, \binits{X.}}:
Deep learning for real-time atari game play using offline monte-carlo tree search planning.
Advances in neural information processing systems
\textbf{27}
(2014)
\end{botherref}
\endbibitem

\bibitem[\protect\citeauthoryear{Silver et~al.}{2016}]{silver2016mastering}
\begin{barticle}
\bauthor{\bsnm{Silver}, \binits{D.}},
\bauthor{\bsnm{Huang}, \binits{A.}},
\bauthor{\bsnm{Maddison}, \binits{C.J.}},
\bauthor{\bsnm{Guez}, \binits{A.}},
\bauthor{\bsnm{Sifre}, \binits{L.}},
\bauthor{\bsnm{Van Den~Driessche}, \binits{G.}},
\bauthor{\bsnm{Schrittwieser}, \binits{J.}},
\bauthor{\bsnm{Antonoglou}, \binits{I.}},
\bauthor{\bsnm{Panneershelvam}, \binits{V.}},
\bauthor{\bsnm{Lanctot}, \binits{M.}}, \betal:
\batitle{Mastering the game of go with deep neural networks and tree search}.
\bjtitle{nature}
\bvolume{529}(\bissue{7587}),
\bfpage{484}--\blpage{489}
(\byear{2016})
\end{barticle}
\endbibitem

\bibitem[\protect\citeauthoryear{Snell et~al.}{2024}]{snell2024scaling}
\begin{botherref}
\oauthor{\bsnm{Snell}, \binits{C.}},
\oauthor{\bsnm{Lee}, \binits{J.}},
\oauthor{\bsnm{Xu}, \binits{K.}},
\oauthor{\bsnm{Kumar}, \binits{A.}}:
Scaling llm test-time compute optimally can be more effective than scaling model parameters.
arXiv preprint arXiv:2408.03314
(2024)
\end{botherref}
\endbibitem

\bibitem[\protect\citeauthoryear{Feng et~al.}{2023}]{feng2023alphazero}
\begin{botherref}
\oauthor{\bsnm{Feng}, \binits{X.}},
\oauthor{\bsnm{Wan}, \binits{Z.}},
\oauthor{\bsnm{Wen}, \binits{M.}},
\oauthor{\bsnm{McAleer}, \binits{S.M.}},
\oauthor{\bsnm{Wen}, \binits{Y.}},
\oauthor{\bsnm{Zhang}, \binits{W.}},
\oauthor{\bsnm{Wang}, \binits{J.}}:
Alphazero-like tree-search can guide large language model decoding and training.
arXiv preprint arXiv:2309.17179
(2023)
\end{botherref}
\endbibitem

\bibitem[\protect\citeauthoryear{Trinh et~al.}{2024}]{trinh2024solving}
\begin{barticle}
\bauthor{\bsnm{Trinh}, \binits{T.H.}},
\bauthor{\bsnm{Wu}, \binits{Y.}},
\bauthor{\bsnm{Le}, \binits{Q.V.}},
\bauthor{\bsnm{He}, \binits{H.}},
\bauthor{\bsnm{Luong}, \binits{T.}}:
\batitle{Solving olympiad geometry without human demonstrations}.
\bjtitle{Nature}
\bvolume{625}(\bissue{7995}),
\bfpage{476}--\blpage{482}
(\byear{2024})
\end{barticle}
\endbibitem

\bibitem[\protect\citeauthoryear{Xin et~al.}{2024}]{xin2024deepseek}
\begin{botherref}
\oauthor{\bsnm{Xin}, \binits{H.}},
\oauthor{\bsnm{Ren}, \binits{Z.}},
\oauthor{\bsnm{Song}, \binits{J.}},
\oauthor{\bsnm{Shao}, \binits{Z.}},
\oauthor{\bsnm{Zhao}, \binits{W.}},
\oauthor{\bsnm{Wang}, \binits{H.}},
\oauthor{\bsnm{Liu}, \binits{B.}},
\oauthor{\bsnm{Zhang}, \binits{L.}},
\oauthor{\bsnm{Lu}, \binits{X.}},
\oauthor{\bsnm{Du}, \binits{Q.}}, et al.:
Deepseek-prover-v1. 5: Harnessing proof assistant feedback for reinforcement learning and monte-carlo tree search.
arXiv preprint arXiv:2408.08152
(2024)
\end{botherref}
\endbibitem

\bibitem[\protect\citeauthoryear{Kumar et~al.}{2024}]{kumar2024training}
\begin{botherref}
\oauthor{\bsnm{Kumar}, \binits{A.}},
\oauthor{\bsnm{Zhuang}, \binits{V.}},
\oauthor{\bsnm{Agarwal}, \binits{R.}},
\oauthor{\bsnm{Su}, \binits{Y.}},
\oauthor{\bsnm{Co-Reyes}, \binits{J.D.}},
\oauthor{\bsnm{Singh}, \binits{A.}},
\oauthor{\bsnm{Baumli}, \binits{K.}},
\oauthor{\bsnm{Iqbal}, \binits{S.}},
\oauthor{\bsnm{Bishop}, \binits{C.}},
\oauthor{\bsnm{Roelofs}, \binits{R.}}, et al.:
Training language models to self-correct via reinforcement learning.
arXiv preprint arXiv:2409.12917
(2024)
\end{botherref}
\endbibitem

\bibitem[\protect\citeauthoryear{Wang et~al.}{2023}]{wang2023math}
\begin{botherref}
\oauthor{\bsnm{Wang}, \binits{P.}},
\oauthor{\bsnm{Li}, \binits{L.}},
\oauthor{\bsnm{Shao}, \binits{Z.}},
\oauthor{\bsnm{Xu}, \binits{R.}},
\oauthor{\bsnm{Dai}, \binits{D.}},
\oauthor{\bsnm{Li}, \binits{Y.}},
\oauthor{\bsnm{Chen}, \binits{D.}},
\oauthor{\bsnm{Wu}, \binits{Y.}},
\oauthor{\bsnm{Sui}, \binits{Z.}}:
Math-shepherd: A label-free step-by-step verifier for llms in mathematical reasoning.
arXiv preprint arXiv:2312.08935
(2023)
\end{botherref}
\endbibitem

\bibitem[\protect\citeauthoryear{Lightman et~al.}{2023}]{lightman2023let}
\begin{botherref}
\oauthor{\bsnm{Lightman}, \binits{H.}},
\oauthor{\bsnm{Kosaraju}, \binits{V.}},
\oauthor{\bsnm{Burda}, \binits{Y.}},
\oauthor{\bsnm{Edwards}, \binits{H.}},
\oauthor{\bsnm{Baker}, \binits{B.}},
\oauthor{\bsnm{Lee}, \binits{T.}},
\oauthor{\bsnm{Leike}, \binits{J.}},
\oauthor{\bsnm{Schulman}, \binits{J.}},
\oauthor{\bsnm{Sutskever}, \binits{I.}},
\oauthor{\bsnm{Cobbe}, \binits{K.}}:
Let's verify step by step.
arXiv preprint arXiv:2305.20050
(2023)
\end{botherref}
\endbibitem

\bibitem[\protect\citeauthoryear{Uesato et~al.}{2022}]{uesato2022solving}
\begin{botherref}
\oauthor{\bsnm{Uesato}, \binits{J.}},
\oauthor{\bsnm{Kushman}, \binits{N.}},
\oauthor{\bsnm{Kumar}, \binits{R.}},
\oauthor{\bsnm{Song}, \binits{F.}},
\oauthor{\bsnm{Siegel}, \binits{N.}},
\oauthor{\bsnm{Wang}, \binits{L.}},
\oauthor{\bsnm{Creswell}, \binits{A.}},
\oauthor{\bsnm{Irving}, \binits{G.}},
\oauthor{\bsnm{Higgins}, \binits{I.}}:
Solving math word problems with process-and outcome-based feedback.
arXiv preprint arXiv:2211.14275
(2022)
\end{botherref}
\endbibitem

\bibitem[\protect\citeauthoryear{Zheng et~al.}{2015}]{zheng2015conditional}
\begin{bchapter}
\bauthor{\bsnm{Zheng}, \binits{S.}},
\bauthor{\bsnm{Jayasumana}, \binits{S.}},
\bauthor{\bsnm{Romera-Paredes}, \binits{B.}},
\bauthor{\bsnm{Vineet}, \binits{V.}},
\bauthor{\bsnm{Su}, \binits{Z.}},
\bauthor{\bsnm{Du}, \binits{D.}},
\bauthor{\bsnm{Huang}, \binits{C.}},
\bauthor{\bsnm{Torr}, \binits{P.H.}}:
\bctitle{Conditional random fields as recurrent neural networks}.
In: \bbtitle{Proceedings of the IEEE International Conference on Computer Vision},
pp. \bfpage{1529}--\blpage{1537}
(\byear{2015})
\end{bchapter}
\endbibitem

\bibitem[\protect\citeauthoryear{Ho et~al.}{2020}]{ho2020denoising}
\begin{barticle}
\bauthor{\bsnm{Ho}, \binits{J.}},
\bauthor{\bsnm{Jain}, \binits{A.}},
\bauthor{\bsnm{Abbeel}, \binits{P.}}:
\batitle{Denoising diffusion probabilistic models}.
\bjtitle{Advances in neural information processing systems}
\bvolume{33},
\bfpage{6840}--\blpage{6851}
(\byear{2020})
\end{barticle}
\endbibitem

\bibitem[\protect\citeauthoryear{Dvornik et~al.}{2020}]{dvornik2020selecting}
\begin{bchapter}
\bauthor{\bsnm{Dvornik}, \binits{N.}},
\bauthor{\bsnm{Schmid}, \binits{C.}},
\bauthor{\bsnm{Mairal}, \binits{J.}}:
\bctitle{Selecting relevant features from a multi-domain representation for few-shot classification}.
In: \bbtitle{European Conference on Computer Vision},
pp. \bfpage{769}--\blpage{786}
(\byear{2020}).
\bcomment{Springer}
\end{bchapter}
\endbibitem

\bibitem[\protect\citeauthoryear{Wang et~al.}{2025}]{wang2025hierarchical}
\begin{botherref}
\oauthor{\bsnm{Wang}, \binits{G.}},
\oauthor{\bsnm{Li}, \binits{J.}},
\oauthor{\bsnm{Sun}, \binits{Y.}},
\oauthor{\bsnm{Chen}, \binits{X.}},
\oauthor{\bsnm{Liu}, \binits{C.}},
\oauthor{\bsnm{Wu}, \binits{Y.}},
\oauthor{\bsnm{Lu}, \binits{M.}},
\oauthor{\bsnm{Song}, \binits{S.}},
\oauthor{\bsnm{Yadkori}, \binits{Y.A.}}:
Hierarchical reasoning model.
arXiv preprint arXiv:2506.21734
(2025)
\end{botherref}
\endbibitem

\bibitem[\protect\citeauthoryear{Jolicoeur-Martineau}{2025}]{jolicoeur2025less}
\begin{botherref}
\oauthor{\bsnm{Jolicoeur-Martineau}, \binits{A.}}:
Less is more: Recursive reasoning with tiny networks.
arXiv preprint arXiv:2510.04871
(2025)
\end{botherref}
\endbibitem

\bibitem[\protect\citeauthoryear{Weston and Sukhbaatar}{2023}]{weston2023system}
\begin{botherref}
\oauthor{\bsnm{Weston}, \binits{J.}},
\oauthor{\bsnm{Sukhbaatar}, \binits{S.}}:
System 2 attention (is something you might need too).
arXiv preprint arXiv:2311.11829
(2023)
\end{botherref}
\endbibitem

\bibitem[\protect\citeauthoryear{Cormen et~al.}{2022}]{cormen2022introduction}
\begin{bbook}
\bauthor{\bsnm{Cormen}, \binits{T.H.}},
\bauthor{\bsnm{Leiserson}, \binits{C.E.}},
\bauthor{\bsnm{Rivest}, \binits{R.L.}},
\bauthor{\bsnm{Stein}, \binits{C.}}:
\bbtitle{Introduction to Algorithms}.
(\byear{2022})
\end{bbook}
\endbibitem

\bibitem[\protect\citeauthoryear{John et~al.}{1994}]{john1994irrelevant}
\begin{bchapter}
\bauthor{\bsnm{John}, \binits{G.H.}},
\bauthor{\bsnm{Kohavi}, \binits{R.}},
\bauthor{\bsnm{Pfleger}, \binits{K.}}:
\bctitle{Irrelevant features and the subset selection problem}.
In: \bbtitle{Machine Learning Proceedings 1994},
pp. \bfpage{121}--\blpage{129}.
(\byear{1994})
\end{bchapter}
\endbibitem

\bibitem[\protect\citeauthoryear{{\v{S}}uch et~al.}{2019}]{vsuch2019neural}
\begin{botherref}
\oauthor{\bsnm{{\v{S}}uch}, \binits{O.}},
\oauthor{\bsnm{Kont{\v{s}}ek}, \binits{M.}},
\oauthor{\bsnm{Tinajov{\'a}}, \binits{A.}}:
Neural pairwise classification models created by ignoring irrelevant alternatives.
ITAT
(2019)
\end{botherref}
\endbibitem

\bibitem[\protect\citeauthoryear{Laakom et~al.}{2021}]{laakom2021learning}
\begin{botherref}
\oauthor{\bsnm{Laakom}, \binits{F.}},
\oauthor{\bsnm{Chumachenko}, \binits{K.}},
\oauthor{\bsnm{Raitoharju}, \binits{J.}},
\oauthor{\bsnm{Iosifidis}, \binits{A.}},
\oauthor{\bsnm{Gabbouj}, \binits{M.}}:
Learning to ignore: rethinking attention in cnns.
arXiv preprint arXiv:2111.05684
(2021)
\end{botherref}
\endbibitem

\bibitem[\protect\citeauthoryear{Vaswani}{2017}]{vaswani2017attention}
\begin{botherref}
\oauthor{\bsnm{Vaswani}, \binits{A.}}:
Attention is all you need.
Advances in Neural Information Processing Systems
(2017)
\end{botherref}
\endbibitem

\bibitem[\protect\citeauthoryear{Ronneberger et~al.}{2015}]{ronneberger2015u}
\begin{bchapter}
\bauthor{\bsnm{Ronneberger}, \binits{O.}},
\bauthor{\bsnm{Fischer}, \binits{P.}},
\bauthor{\bsnm{Brox}, \binits{T.}}:
\bctitle{U-net: Convolutional networks for biomedical image segmentation}.
In: \bbtitle{Medical Image Computing and Computer-assisted intervention--MICCAI 2015: 18th International Conference, Munich, Germany, October 5-9, 2015, Proceedings, Part III 18},
pp. \bfpage{234}--\blpage{241}
(\byear{2015}).
\bcomment{Springer}
\end{bchapter}
\endbibitem

\bibitem[\protect\citeauthoryear{Gao et~al.}{2022a}]{gao2022large}
\begin{barticle}
\bauthor{\bsnm{Gao}, \binits{S.}},
\bauthor{\bsnm{Li}, \binits{Z.-Y.}},
\bauthor{\bsnm{Yang}, \binits{M.-H.}},
\bauthor{\bsnm{Cheng}, \binits{M.-M.}},
\bauthor{\bsnm{Han}, \binits{J.}},
\bauthor{\bsnm{Torr}, \binits{P.}}:
\batitle{Large-scale unsupervised semantic segmentation}.
\bjtitle{IEEE transactions on pattern analysis and machine intelligence}
\bvolume{45}(\bissue{6}),
\bfpage{7457}--\blpage{7476}
(\byear{2022})
\end{barticle}
\endbibitem

\bibitem[\protect\citeauthoryear{Gao et~al.}{2022b}]{gao2022towards}
\begin{botherref}
\oauthor{\bsnm{Gao}, \binits{S.}},
\oauthor{\bsnm{Zhou}, \binits{P.}},
\oauthor{\bsnm{Cheng}, \binits{M.-M.}},
\oauthor{\bsnm{Yan}, \binits{S.}}:
Towards sustainable self-supervised learning.
arXiv preprint arXiv:2210.11016
(2022)
\end{botherref}
\endbibitem

\bibitem[\protect\citeauthoryear{Ahmed et~al.}{2017}]{ahmed2017diagnostic}
\begin{barticle}
\bauthor{\bsnm{Ahmed}, \binits{H.U.}},
\bauthor{\bsnm{Bosaily}, \binits{A.E.-S.}},
\bauthor{\bsnm{Brown}, \binits{L.C.}},
\bauthor{\bsnm{Gabe}, \binits{R.}},
\bauthor{\bsnm{Kaplan}, \binits{R.}},
\bauthor{\bsnm{Parmar}, \binits{M.K.}},
\bauthor{\bsnm{Collaco-Moraes}, \binits{Y.}},
\bauthor{\bsnm{Ward}, \binits{K.}},
\bauthor{\bsnm{Hindley}, \binits{R.G.}},
\bauthor{\bsnm{Freeman}, \binits{A.}}, \betal:
\batitle{Diagnostic accuracy of multi-parametric mri and trus biopsy in prostate cancer (promis): a paired validating confirmatory study}.
\bjtitle{The Lancet}
\bvolume{389}(\bissue{10071}),
\bfpage{815}--\blpage{822}
(\byear{2017})
\end{barticle}
\endbibitem

\bibitem[\protect\citeauthoryear{Krajewski and Pedrosa}{2018}]{krajewski2018imaging}
\begin{barticle}
\bauthor{\bsnm{Krajewski}, \binits{K.M.}},
\bauthor{\bsnm{Pedrosa}, \binits{I.}}:
\batitle{Imaging advances in the management of kidney cancer}.
\bjtitle{Journal of Clinical Oncology}
\bvolume{36}(\bissue{36}),
\bfpage{3582}--\blpage{3590}
(\byear{2018})
\end{barticle}
\endbibitem

\bibitem[\protect\citeauthoryear{Oliva and Saini}{2004}]{oliva2004liver}
\begin{barticle}
\bauthor{\bsnm{Oliva}, \binits{M.R.}},
\bauthor{\bsnm{Saini}, \binits{S.}}:
\batitle{Liver cancer imaging: role of ct, mri, us and pet}.
\bjtitle{Cancer imaging}
\bvolume{4}(\bissue{Spec No A}),
\bfpage{42}
(\byear{2004})
\end{barticle}
\endbibitem

\bibitem[\protect\citeauthoryear{Hunter et~al.}{2022}]{hunter2022role}
\begin{barticle}
\bauthor{\bsnm{Hunter}, \binits{B.}},
\bauthor{\bsnm{Hindocha}, \binits{S.}},
\bauthor{\bsnm{Lee}, \binits{R.W.}}:
\batitle{The role of artificial intelligence in early cancer diagnosis}.
\bjtitle{Cancers}
\bvolume{14}(\bissue{6}),
\bfpage{1524}
(\byear{2022})
\end{barticle}
\endbibitem

\bibitem[\protect\citeauthoryear{Sprague et~al.}{2016}]{sprague2016variation}
\begin{barticle}
\bauthor{\bsnm{Sprague}, \binits{B.L.}},
\bauthor{\bsnm{Conant}, \binits{E.F.}},
\bauthor{\bsnm{Onega}, \binits{T.}},
\bauthor{\bsnm{Garcia}, \binits{M.P.}},
\bauthor{\bsnm{Beaber}, \binits{E.F.}},
\bauthor{\bsnm{Herschorn}, \binits{S.D.}},
\bauthor{\bsnm{Lehman}, \binits{C.D.}},
\bauthor{\bsnm{Tosteson}, \binits{A.N.}},
\bauthor{\bsnm{Lacson}, \binits{R.}},
\bauthor{\bsnm{Schnall}, \binits{M.D.}}, \betal:
\batitle{Variation in mammographic breast density assessments among radiologists in clinical practice: a multicenter observational study}.
\bjtitle{Annals of internal medicine}
\bvolume{165}(\bissue{7}),
\bfpage{457}--\blpage{464}
(\byear{2016})
\end{barticle}
\endbibitem

\bibitem[\protect\citeauthoryear{Vleugels et~al.}{2020}]{vleugels2020suboptimal}
\begin{barticle}
\bauthor{\bsnm{Vleugels}, \binits{J.L.}},
\bauthor{\bsnm{Koens}, \binits{L.}},
\bauthor{\bsnm{Dijkgraaf}, \binits{M.G.}},
\bauthor{\bsnm{Houwen}, \binits{B.}},
\bauthor{\bsnm{Hazewinkel}, \binits{Y.}},
\bauthor{\bsnm{Fockens}, \binits{P.}},
\bauthor{\bsnm{Dekker}, \binits{E.}}:
\batitle{Suboptimal endoscopic cancer recognition in colorectal lesions in a national bowel screening programme}.
\bjtitle{Gut}
\bvolume{69}(\bissue{6}),
\bfpage{977}--\blpage{980}
(\byear{2020})
\end{barticle}
\endbibitem

\bibitem[\protect\citeauthoryear{Czolbe et~al.}{2021}]{czolbe2021segmentation}
\begin{bchapter}
\bauthor{\bsnm{Czolbe}, \binits{S.}},
\bauthor{\bsnm{Arnavaz}, \binits{K.}},
\bauthor{\bsnm{Krause}, \binits{O.}},
\bauthor{\bsnm{Feragen}, \binits{A.}}:
\bctitle{Is segmentation uncertainty useful?}
In: \bbtitle{Information Processing in Medical Imaging: 27th International Conference, IPMI 2021, Virtual Event, June 28--June 30, 2021, Proceedings 27},
pp. \bfpage{715}--\blpage{726}
(\byear{2021}).
\bcomment{Springer}
\end{bchapter}
\endbibitem

\bibitem[\protect\citeauthoryear{Saeed et~al.}{2024}]{saeed2024active}
\begin{barticle}
\bauthor{\bsnm{Saeed}, \binits{S.U.}},
\bauthor{\bsnm{Ramalhinho}, \binits{J.}},
\bauthor{\bsnm{Pinnock}, \binits{M.}},
\bauthor{\bsnm{Shen}, \binits{Z.}},
\bauthor{\bsnm{Fu}, \binits{Y.}},
\bauthor{\bsnm{Monta{\~n}a-Brown}, \binits{N.}},
\bauthor{\bsnm{Bonmati}, \binits{E.}},
\bauthor{\bsnm{Barratt}, \binits{D.C.}},
\bauthor{\bsnm{Pereira}, \binits{S.P.}},
\bauthor{\bsnm{Davidson}, \binits{B.}}, \betal:
\batitle{Active learning using adaptable task-based prioritisation}.
\bjtitle{Medical Image Analysis}
\bvolume{95},
\bfpage{103181}
(\byear{2024})
\end{barticle}
\endbibitem

\bibitem[\protect\citeauthoryear{Saeed et~al.}{2022a}]{saeed2022image}
\begin{botherref}
\oauthor{\bsnm{Saeed}, \binits{S.U.}},
\oauthor{\bsnm{Yan}, \binits{W.}},
\oauthor{\bsnm{Fu}, \binits{Y.}},
\oauthor{\bsnm{Giganti}, \binits{F.}},
\oauthor{\bsnm{Yang}, \binits{Q.}},
\oauthor{\bsnm{Baum}, \binits{Z.}},
\oauthor{\bsnm{Rusu}, \binits{M.}},
\oauthor{\bsnm{Fan}, \binits{R.E.}},
\oauthor{\bsnm{Sonn}, \binits{G.A.}},
\oauthor{\bsnm{Emberton}, \binits{M.}}, et al.:
Image quality assessment by overlapping task-specific and task-agnostic measures: application to prostate multiparametric mr images for cancer segmentation.
arXiv preprint arXiv:2202.09798
(2022)
\end{botherref}
\endbibitem

\bibitem[\protect\citeauthoryear{Saeed et~al.}{2022b}]{saeed2022image_media}
\begin{barticle}
\bauthor{\bsnm{Saeed}, \binits{S.U.}},
\bauthor{\bsnm{Fu}, \binits{Y.}},
\bauthor{\bsnm{Stavrinides}, \binits{V.}},
\bauthor{\bsnm{Baum}, \binits{Z.M.}},
\bauthor{\bsnm{Yang}, \binits{Q.}},
\bauthor{\bsnm{Rusu}, \binits{M.}},
\bauthor{\bsnm{Fan}, \binits{R.E.}},
\bauthor{\bsnm{Sonn}, \binits{G.A.}},
\bauthor{\bsnm{Noble}, \binits{J.A.}},
\bauthor{\bsnm{Barratt}, \binits{D.C.}}, \betal:
\batitle{Image quality assessment for machine learning tasks using meta-reinforcement learning}.
\bjtitle{Medical Image Analysis}
\bvolume{78},
\bfpage{102427}
(\byear{2022})
\end{barticle}
\endbibitem

\bibitem[\protect\citeauthoryear{Yan et~al.}{2022}]{yan2022impact}
\begin{bchapter}
\bauthor{\bsnm{Yan}, \binits{W.}},
\bauthor{\bsnm{Yang}, \binits{Q.}},
\bauthor{\bsnm{Syer}, \binits{T.}},
\bauthor{\bsnm{Min}, \binits{Z.}},
\bauthor{\bsnm{Punwani}, \binits{S.}},
\bauthor{\bsnm{Emberton}, \binits{M.}},
\bauthor{\bsnm{Barratt}, \binits{D.}},
\bauthor{\bsnm{Chiu}, \binits{B.}},
\bauthor{\bsnm{Hu}, \binits{Y.}}:
\bctitle{The impact of using voxel-level segmentation metrics on evaluating multifocal prostate cancer localisation}.
In: \bbtitle{International Workshop on Applications of Medical AI},
pp. \bfpage{128}--\blpage{138}
(\byear{2022}).
\bcomment{Springer}
\end{bchapter}
\endbibitem

\bibitem[\protect\citeauthoryear{Pocius et~al.}{2024}]{pocius2024weakly}
\begin{botherref}
\oauthor{\bsnm{Pocius}, \binits{M.}},
\oauthor{\bsnm{Yan}, \binits{W.}},
\oauthor{\bsnm{Barratt}, \binits{D.C.}},
\oauthor{\bsnm{Emberton}, \binits{M.}},
\oauthor{\bsnm{Clarkson}, \binits{M.J.}},
\oauthor{\bsnm{Hu}, \binits{Y.}},
\oauthor{\bsnm{Saeed}, \binits{S.U.}}:
Weakly supervised localisation of prostate cancer using reinforcement learning for bi-parametric mr images.
arXiv preprint arXiv:2402.13778
(2024)
\end{botherref}
\endbibitem

\bibitem[\protect\citeauthoryear{Liu et~al.}{2023}]{liu2023clip}
\begin{bchapter}
\bauthor{\bsnm{Liu}, \binits{J.}},
\bauthor{\bsnm{Zhang}, \binits{Y.}},
\bauthor{\bsnm{Chen}, \binits{J.-N.}},
\bauthor{\bsnm{Xiao}, \binits{J.}},
\bauthor{\bsnm{Lu}, \binits{Y.}},
\bauthor{\bsnm{A~Landman}, \binits{B.}},
\bauthor{\bsnm{Yuan}, \binits{Y.}},
\bauthor{\bsnm{Yuille}, \binits{A.}},
\bauthor{\bsnm{Tang}, \binits{Y.}},
\bauthor{\bsnm{Zhou}, \binits{Z.}}:
\bctitle{Clip-driven universal model for organ segmentation and tumor detection}.
In: \bbtitle{Proceedings of the IEEE/CVF International Conference on Computer Vision},
pp. \bfpage{21152}--\blpage{21164}
(\byear{2023})
\end{bchapter}
\endbibitem

\bibitem[\protect\citeauthoryear{Myronenko et~al.}{2023}]{myronenko2023automated}
\begin{bchapter}
\bauthor{\bsnm{Myronenko}, \binits{A.}},
\bauthor{\bsnm{Yang}, \binits{D.}},
\bauthor{\bsnm{He}, \binits{Y.}},
\bauthor{\bsnm{Xu}, \binits{D.}}:
\bctitle{Automated 3d segmentation of kidneys and tumors in miccai kits 2023 challenge}.
In: \bbtitle{International Challenge on Kidney and Kidney Tumor Segmentation},
pp. \bfpage{1}--\blpage{7}.
(\byear{2023})
\end{bchapter}
\endbibitem

\bibitem[\protect\citeauthoryear{Heller et~al.}{2019}]{heller2019kits19}
\begin{botherref}
\oauthor{\bsnm{Heller}, \binits{N.}},
\oauthor{\bsnm{Sathianathen}, \binits{N.}},
\oauthor{\bsnm{Kalapara}, \binits{A.}},
\oauthor{\bsnm{Walczak}, \binits{E.}},
\oauthor{\bsnm{Moore}, \binits{K.}},
\oauthor{\bsnm{Kaluzniak}, \binits{H.}},
\oauthor{\bsnm{Rosenberg}, \binits{J.}},
\oauthor{\bsnm{Blake}, \binits{P.}},
\oauthor{\bsnm{Rengel}, \binits{Z.}},
\oauthor{\bsnm{Oestreich}, \binits{M.}}, et al.:
The kits19 challenge data: 300 kidney tumor cases with clinical context, ct semantic segmentations, and surgical outcomes.
arXiv preprint arXiv:1904.00445
(2019)
\end{botherref}
\endbibitem

\bibitem[\protect\citeauthoryear{Veiga-Canuto et~al.}{2022}]{veiga2022comparative}
\begin{barticle}
\bauthor{\bsnm{Veiga-Canuto}, \binits{D.}},
\bauthor{\bsnm{Cerd{\`a}-Alberich}, \binits{L.}},
\bauthor{\bsnm{Sang{\"u}esa~Nebot}, \binits{C.}},
\bauthor{\bsnm{Heras}, \binits{B.}},
\bauthor{\bsnm{P{\"o}tschger}, \binits{U.}},
\bauthor{\bsnm{Gabelloni}, \binits{M.}},
\bauthor{\bsnm{Carot~Sierra}, \binits{J.M.}},
\bauthor{\bsnm{Taschner-Mandl}, \binits{S.}},
\bauthor{\bsnm{D{\"u}ster}, \binits{V.}},
\bauthor{\bsnm{Ca{\~n}ete}, \binits{A.}}, \betal:
\batitle{Comparative multicentric evaluation of inter-observer variability in manual and automatic segmentation of neuroblastic tumors in magnetic resonance images}.
\bjtitle{Cancers}
\bvolume{14}(\bissue{15}),
\bfpage{3648}
(\byear{2022})
\end{barticle}
\endbibitem

\bibitem[\protect\citeauthoryear{Oquab et~al.}{2023}]{oquab2023dinov2}
\begin{botherref}
\oauthor{\bsnm{Oquab}, \binits{M.}},
\oauthor{\bsnm{Darcet}, \binits{T.}},
\oauthor{\bsnm{Moutakanni}, \binits{T.}},
\oauthor{\bsnm{Vo}, \binits{H.}},
\oauthor{\bsnm{Szafraniec}, \binits{M.}},
\oauthor{\bsnm{Khalidov}, \binits{V.}},
\oauthor{\bsnm{Fernandez}, \binits{P.}},
\oauthor{\bsnm{Haziza}, \binits{D.}},
\oauthor{\bsnm{Massa}, \binits{F.}},
\oauthor{\bsnm{El-Nouby}, \binits{A.}}, et al.:
Dinov2: Learning robust visual features without supervision.
arXiv preprint arXiv:2304.07193
(2023)
\end{botherref}
\endbibitem

\bibitem[\protect\citeauthoryear{Butoi et~al.}{2023}]{butoi2023universeg}
\begin{bchapter}
\bauthor{\bsnm{Butoi}, \binits{V.I.}},
\bauthor{\bsnm{Ortiz}, \binits{J.J.G.}},
\bauthor{\bsnm{Ma}, \binits{T.}},
\bauthor{\bsnm{Sabuncu}, \binits{M.R.}},
\bauthor{\bsnm{Guttag}, \binits{J.}},
\bauthor{\bsnm{Dalca}, \binits{A.V.}}:
\bctitle{Universeg: Universal medical image segmentation}.
In: \bbtitle{Proceedings of the IEEE/CVF International Conference on Computer Vision},
pp. \bfpage{21438}--\blpage{21451}
(\byear{2023})
\end{bchapter}
\endbibitem

\bibitem[\protect\citeauthoryear{Goodfellow et~al.}{2020}]{goodfellow2020generative}
\begin{barticle}
\bauthor{\bsnm{Goodfellow}, \binits{I.}},
\bauthor{\bsnm{Pouget-Abadie}, \binits{J.}},
\bauthor{\bsnm{Mirza}, \binits{M.}},
\bauthor{\bsnm{Xu}, \binits{B.}},
\bauthor{\bsnm{Warde-Farley}, \binits{D.}},
\bauthor{\bsnm{Ozair}, \binits{S.}},
\bauthor{\bsnm{Courville}, \binits{A.}},
\bauthor{\bsnm{Bengio}, \binits{Y.}}:
\batitle{Generative adversarial networks}.
\bjtitle{Communications of the ACM}
\bvolume{63}(\bissue{11}),
\bfpage{139}--\blpage{144}
(\byear{2020})
\end{barticle}
\endbibitem

\bibitem[\protect\citeauthoryear{Yun et~al.}{2013}]{yun2013exploring}
\begin{barticle}
\bauthor{\bsnm{Yun}, \binits{K.}},
\bauthor{\bsnm{Peng}, \binits{Y.}},
\bauthor{\bsnm{Samaras}, \binits{D.}},
\bauthor{\bsnm{Zelinsky}, \binits{G.J.}},
\bauthor{\bsnm{Berg}, \binits{T.L.}}:
\batitle{Exploring the role of gaze behavior and object detection in scene understanding}.
\bjtitle{Frontiers in psychology}
\bvolume{4},
\bfpage{917}
(\byear{2013})
\end{barticle}
\endbibitem

\bibitem[\protect\citeauthoryear{Nichol et~al.}{2018}]{nichol2018first}
\begin{botherref}
\oauthor{\bsnm{Nichol}, \binits{A.}},
\oauthor{\bsnm{Achiam}, \binits{J.}},
\oauthor{\bsnm{Schulman}, \binits{J.}}:
On first-order meta-learning algorithms.
arXiv preprint arXiv:1803.02999
(2018)
\end{botherref}
\endbibitem

\bibitem[\protect\citeauthoryear{Silver et~al.}{2018}]{silver2018general}
\begin{barticle}
\bauthor{\bsnm{Silver}, \binits{D.}},
\bauthor{\bsnm{Hubert}, \binits{T.}},
\bauthor{\bsnm{Schrittwieser}, \binits{J.}},
\bauthor{\bsnm{Antonoglou}, \binits{I.}},
\bauthor{\bsnm{Lai}, \binits{M.}},
\bauthor{\bsnm{Guez}, \binits{A.}},
\bauthor{\bsnm{Lanctot}, \binits{M.}},
\bauthor{\bsnm{Sifre}, \binits{L.}},
\bauthor{\bsnm{Kumaran}, \binits{D.}},
\bauthor{\bsnm{Graepel}, \binits{T.}}, \betal:
\batitle{A general reinforcement learning algorithm that masters chess, shogi, and go through self-play}.
\bjtitle{Science}
\bvolume{362}(\bissue{6419}),
\bfpage{1140}--\blpage{1144}
(\byear{2018})
\end{barticle}
\endbibitem

\bibitem[\protect\citeauthoryear{Saeed et~al.}{2024}]{saeed2024competing}
\begin{botherref}
\oauthor{\bsnm{Saeed}, \binits{S.U.}},
\oauthor{\bsnm{Huang}, \binits{S.}},
\oauthor{\bsnm{Ramalhinho}, \binits{J.}},
\oauthor{\bsnm{Gayo}, \binits{I.J.}},
\oauthor{\bsnm{Montana-Brown}, \binits{N.}},
\oauthor{\bsnm{Bonmati}, \binits{E.}},
\oauthor{\bsnm{Pereira}, \binits{S.P.}},
\oauthor{\bsnm{Davidson}, \binits{B.}},
\oauthor{\bsnm{Barratt}, \binits{D.C.}},
\oauthor{\bsnm{Clarkson}, \binits{M.J.}}, et al.:
Competing for pixels: a self-play algorithm for weakly-supervised semantic segmentation.
IEEE Transactions on Pattern Analysis and Machine Intelligence
(2024)
\end{botherref}
\endbibitem

\bibitem[\protect\citeauthoryear{Saeed}{2026}]{shaheer_u_saeed_2026_19331690}
\begin{botherref}
\oauthor{\bsnm{Saeed}, \binits{S.U.}}:
s-sd/system-II-vision: Open Source Release.
\doiurl{10.5281/zenodo.19331690} .
\url{https://doi.org/10.5281/zenodo.19331690}
\end{botherref}
\endbibitem



\bibitem[\protect\citeauthoryear{Antonelli et~al.}{2021}]{antonelli_decathalon}
\begin{botherref}
\oauthor{\bsnm{Antonelli}, \binits{M.}},
\oauthor{\bsnm{Reinke}, \binits{A.}},
\oauthor{\bsnm{Bakas}, \binits{S.}},
\oauthor{\bsnm{Farahani}, \binits{K.}},
\oauthor{\bsnm{{AnnetteKopp-Schneider}}},
\oauthor{\bsnm{Landman}, \binits{B.A.}},
\oauthor{\bsnm{Litjens}, \binits{G.}},
\oauthor{\bsnm{Menze}, \binits{B.}},
\oauthor{\bsnm{Ronneberger}, \binits{O.}},
\oauthor{\bsnm{Summers}, \binits{R.M.}},
\oauthor{\bsnm{Ginneken}, \binits{B.}},
\oauthor{\bsnm{Bilello}, \binits{M.}},
\oauthor{\bsnm{Bilic}, \binits{P.}},
\oauthor{\bsnm{Christ}, \binits{P.F.}},
\oauthor{\bsnm{Do}, \binits{R.K.G.}},
\oauthor{\bsnm{Gollub}, \binits{M.J.}},
\oauthor{\bsnm{Heckers}, \binits{S.H.}},
\oauthor{\bsnm{Huisman}, \binits{H.}},
\oauthor{\bsnm{Jarnagin}, \binits{W.R.}},
\oauthor{\bsnm{McHugo}, \binits{M.K.}},
\oauthor{\bsnm{Napel}, \binits{S.}},
\oauthor{\bsnm{Pernicka}, \binits{J.S.G.}},
\oauthor{\bsnm{Rhode}, \binits{K.}},
\oauthor{\bsnm{Tobon-Gomez}, \binits{C.}},
\oauthor{\bsnm{Vorontsov}, \binits{E.}},
\oauthor{\bsnm{Huisman}, \binits{H.}},
\oauthor{\bsnm{Meakin}, \binits{J.A.}},
\oauthor{\bsnm{Ourselin}, \binits{S.}},
\oauthor{\bsnm{Wiesenfarth}, \binits{M.}},
\oauthor{\bsnm{Arbelaez}, \binits{P.}},
\oauthor{\bsnm{Bae}, \binits{B.}},
\oauthor{\bsnm{Chen}, \binits{S.}},
\oauthor{\bsnm{Daza}, \binits{L.}},
\oauthor{\bsnm{Feng}, \binits{J.}},
\oauthor{\bsnm{He}, \binits{B.}},
\oauthor{\bsnm{Isensee}, \binits{F.}},
\oauthor{\bsnm{Ji}, \binits{Y.}},
\oauthor{\bsnm{Jia}, \binits{F.}},
\oauthor{\bsnm{Kim}, \binits{N.}},
\oauthor{\bsnm{Kim}, \binits{I.}},
\oauthor{\bsnm{Merhof}, \binits{D.}},
\oauthor{\bsnm{Pai}, \binits{A.}},
\oauthor{\bsnm{Park}, \binits{B.}},
\oauthor{\bsnm{Perslev}, \binits{M.}},
\oauthor{\bsnm{Rezaiifar}, \binits{R.}},
\oauthor{\bsnm{Rippel}, \binits{O.}},
\oauthor{\bsnm{Sarasua}, \binits{I.}},
\oauthor{\bsnm{Shen}, \binits{W.}},
\oauthor{\bsnm{Son}, \binits{J.}},
\oauthor{\bsnm{Wachinger}, \binits{C.}},
\oauthor{\bsnm{Wang}, \binits{L.}},
\oauthor{\bsnm{Wang}, \binits{Y.}},
\oauthor{\bsnm{Xia}, \binits{Y.}},
\oauthor{\bsnm{Xu}, \binits{D.}},
\oauthor{\bsnm{Xu}, \binits{Z.}},
\oauthor{\bsnm{Zheng}, \binits{Y.}},
\oauthor{\bsnm{Simpson}, \binits{A.L.}},
\oauthor{\bsnm{Maier-Hein}, \binits{L.}},
\oauthor{\bsnm{Cardoso}, \binits{M.J.}}:
The Medical Segmentation Decathlon.
arXiv
(2021)
\end{botherref}
\endbibitem


\bibitem[\protect\citeauthoryear{Heller et~al.}{2021}]{kits}
\begin{barticle}
\bauthor{\bsnm{Heller}, \binits{N.}},
\bauthor{\bsnm{Isensee}, \binits{F.}},
\bauthor{\bsnm{Maier-Hein}, \binits{K.H.}},
\bauthor{\bsnm{Hou}, \binits{X.}},
\bauthor{\bsnm{Xie}, \binits{C.}},
\bauthor{\bsnm{Li}, \binits{F.}},
\bauthor{\bsnm{Nan}, \binits{Y.}},
\bauthor{\bsnm{Mu}, \binits{G.}},
\bauthor{\bsnm{Lin}, \binits{Z.}},
\bauthor{\bsnm{Han}, \binits{M.}}, \betal:
\batitle{The state of the art in kidney and kidney tumor segmentation in contrast-enhanced ct imaging: Results of the kits19 challenge}.
\bjtitle{Medical image analysis}
\bvolume{67},
\bfpage{101821}
(\byear{2021})
\end{barticle}
\endbibitem


\end{thebibliography}

\newpage

\section*{Funding:} S.U.S., Y.H. and V.K. are supported by the International Alliance for Cancer Early Detection, an alliance between Cancer Research UK [EDDAPA-2024/100014] \& [C73666/A31378], Canary Center at Stanford University, the University of Cambridge, OHSU Knight Cancer Institute, University College London and the University of Manchester; M.J.C, Y.H, B.R.D and S.U.S are supported by the EPSRC grant [EP/T029404/1]; and Y.H. and S.U.S. are supported by the National Institute for Health Research University College London Hospitals Biomedical Research Centre.

\section*{Author contributions:}
\noindent {S.U.S.:} Conceptualisation, data curation, formal analysis, funding acquisition, investigation, methodology, software, validation, visualisation, writing (original draft)\\
\noindent {Y.W.:} Validation, visualisation, writing (review and editing)\\
\noindent {V.K.:} Data curation, funding acquisition, resources, supervision, validation, visualisation, writing (review and editing)\\
\noindent {B.R.D.:} Data curation, funding acquisition, resources, supervision, validation, visualisation, writing (review and editing)\\
\noindent {M.J.C.:} Funding acquisition, resources, supervision, validation, visualisation, writing (review and editing)\\
\noindent {Y.H.:} Conceptualisation, data curation, formal analysis, funding acquisition, investigation, methodology, resources, supervision, visualisation, writing (original draft)\\
\noindent {D.C.A:} Investigation, methodology, resources, supervision, visualisation, writing (review and editing)

\section*{Competing interests:}
The authors have no competing interests to declare.

\section*{Figure Legends/Captions} 
\vspace{0.2cm}

\noindent Figure 1: Examples illustrating System I and System II cognitive processes. a) and b) represent problems solved by System I and System II thinking, respectively. c) represents a problem that has transitioned from System II to System I over time. Note: a) and c) are open stock photos available via an institutional Microsoft 365 subscription, and b) is generated using DALLE-2 available via OpenAI ChatGPT under a Plus subscription.

\vspace{0.2cm}

\noindent Figure 2: a) and b) present an overview of the training and evaluation of the proposed System II approach. c) and d) show performance of the System II approach compared with common methods (numbers of samples used for training/ adaptation, from the target task, in brackets). Note: images in a), b) and c) are derived from the public domain MNIST dataset, the original ImageNet images are omitted from d) due to copyright; however, these and corresponding segmentations are available via the online repository: \url{github.com/s-sd/system-II-vision} \cite{shaheer_u_saeed_2026_19331690}.

\vspace{0.2cm}

\noindent Figure 3: Performance of System II thinking compared with deep learning (DL) and foundation models (FM) in five cancer localisation tasks for different organs and image types. The left column shows statistically significant increases in cancer localisation performance compared to DL and FM with equivalent training data (indicated by the numbers in brackets below), and better or at least comparable performance even when DL and FM use substantially more training data. The remaining columns provide qualitative examples demonstrating the improvement as the inference-time compute increases, further discussed in the text.

\vspace{0.2cm}

\noindent Figure 4: System II thinking: a) improves performance with increasing thinking time, compared to supervised deep learning (DL) and foundation models (FM) with fixed inference time (measured in the unit of millions of optimisation iterations); b) outperforms the current best-performing deep learning (DL) methods, using substantially fewer labelled task-specific samples; c) attains peak performance with 8 labelled samples, compared to other methods requiring more than one hundred labelled samples; d) transitions tasks to fast thinking, reducing thinking time as more unlabelled samples are fully refined (or solved) by System II thinking. Performance in plotted lines is the average across 20 samples from the test set. Complete results for all cancer types can be found in the `Comparison to existing methods for cancer localisation' section in the supplementary materials.

\vspace{0.2cm}

\noindent Figure 5: Different learning strategies are compared: deep learning for a single forward pass prediction, meta-learning for adapting to new tasks (akin to fine-tuning pre-trained foundation models), and the proposed System II cognition for allowing inference-time compute. Details of these are described in text.

\end{document}


\date{}
\title{Reasoning in machine vision by learning fast and slow thinking\\Supplementary Material}

\maketitle

\begin{center}
{\large{Shaheer U. Saeed$^{1,2,3,4,5\ast}$, Yipei Wang$^{4,5}$, \\
Veeru Kasivisvanathan$^{6,7}$, Brian R. Davidson$^{8}$,\\
Matthew J. Clarkson$^{4,5}$, Yipeng Hu$^{4,5,9}$, Daniel C. Alexander$^{4,9}$}}    
\end{center}

\vspace{0.5cm}

\begin{center}
{\normalsize{$^{1}$Centre for Bioengineering, Queen Mary University of London, London, UK} \\
\normalsize{$^{2}$Digital Environment Research Institute, Queen Mary University of London, London, UK} \\
\normalsize{$^{3}$School of Engineering and Materials Science, Queen Mary University of London, London, UK} \\
\normalsize{$^{4}$UCL Hawkes Institute, University College London, London, UK}\\
\normalsize{$^{5}$Department of Medical Physics and Biomedical Engineering, University College London, London, UK}\\
\normalsize{$^{6}$Centre for Urology Imaging, Prostate, AI and Surgical Studies (COMPASS) Research Group, Division of Surgery and Interventional Science, University College London, London, UK} \\
\normalsize{$^{7}$Department of Urology, Comprehensive Cancer Center, Medical University of Vienna, Vienna, Austria}\\
\normalsize{$^{8}$Division of Surgery and Interventional Science, University College London, London, UK} \\
\normalsize{$^{9}$Department of Computer Science, University College London, London, UK} \\
\normalsize{$^\ast$Correspondence e-mail: shaheer.saeed@qmul.ac.uk}
}
\end{center}

\section*{Content}

Methodological description\\
Methodological definitions\\
Materials and experimental details\\
Supplementary results\\
Supplementary Figure S1 to S3\\
Algo. S1 to S2\\
Supplementary Table S1 to S9\\
References [1] to [34]\\


\newpage
\subsection*{Methodological description}
\label{sec:method}

Our method emulates human dual-process cognition by integrating two
distinct yet interconnected modules: the System I module, responsible for fast inference, and the System II module, designed for resource-intensive reasoning.

For the System I module, we use adversarial learning \cite{goodfellow2020generative} to train two deep neural networks: the ‘task-predictor’ which predicts solutions to the task (such as segmenting a cat in an image), and the ‘distribution-discriminator’ which decides how likely the solution is to be ‘correct’. During training, the likelihood of the solution being correct is determined by comparing task-predictor solutions to human labels. When these networks are trained across various tasks, the respective capabilities, generating solutions and predicting correctness of these solutions, can be adapted to new tasks. In general, however, the direct prediction from task-predictor performs poorly in unfamiliar or complex scenarios, and therefore resorts to the System II module to reason about and refine solutions.

The System II module not only conducts inference-time computation to refine solutions but also updates the System I module with new insights, reducing reliance on System II for similar tasks, after the one-time inference stage that is compute-intensive. This module consists of two neural networks competing against each other by iteratively hypothesising refinements to solutions (e.g., for segmenting a dog in an image, the refinement may be a change in whether certain pixels belong to the dog or not). The initial solution comes form an adapted task-predictor from the System I module, whilst the likelihood of the refinements being correct in the competing process is predicted by the adapted distribution-discriminator. The initial solution is refined until no further improvement in the likelihood of the solution being correct is possible, at which point the sample is considered `solved'. The solved samples are used to update the System I module, which can then better and much more efficiently predict solutions for similar future tasks. This approach mimics human cognition which commits new scenarios to memory, reducing System II thinking time for future tasks. With sufficient experience, tasks may fully transition to System I, where System II thinking may no longer be required.


We present a mathematical formulation for our method in the following subsections. An overview is presented in Supplementary Figure \ref{fig:method_figure}.

\begin{figure}
    \centering
    \includegraphics[width=0.9\textwidth]{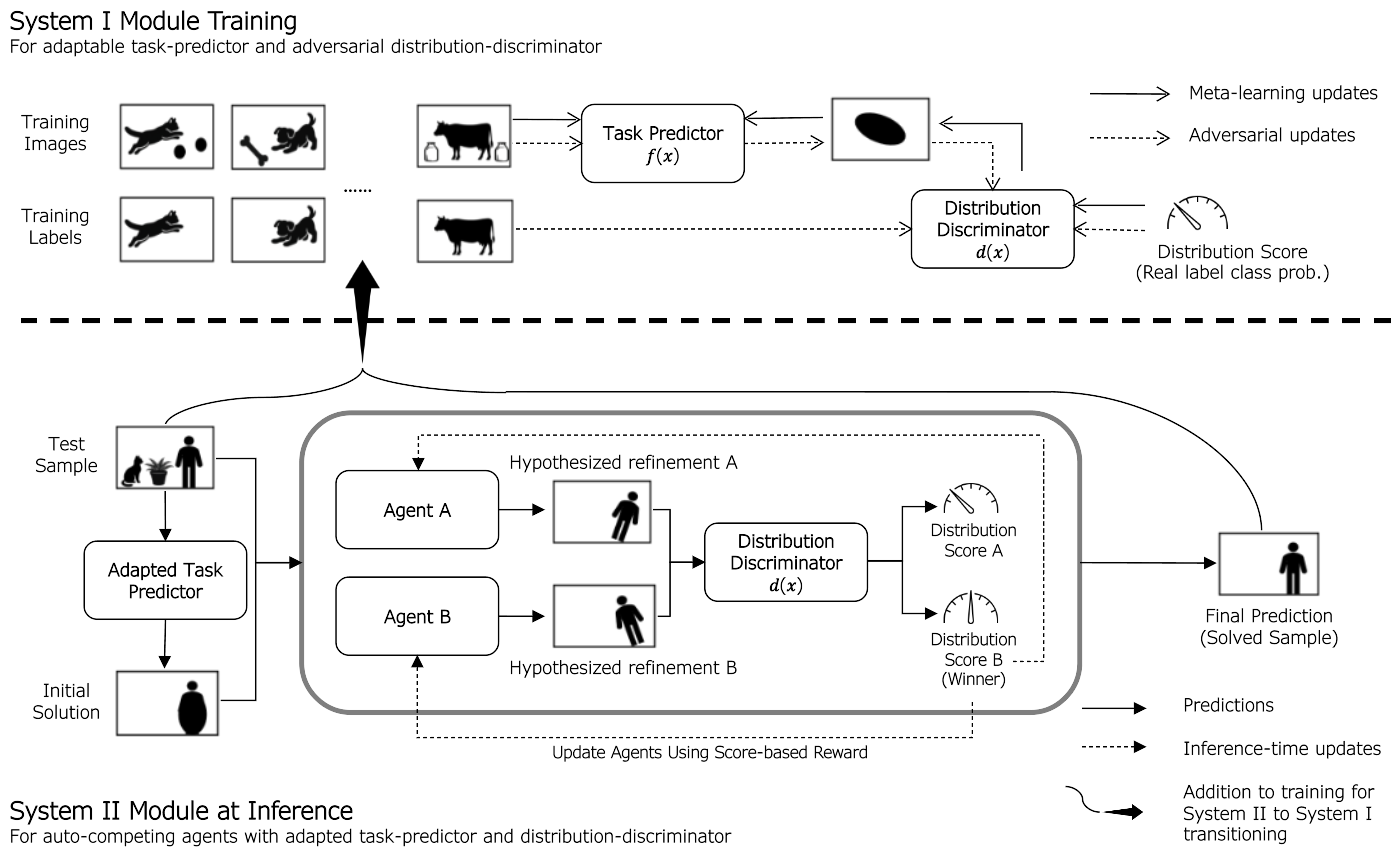}
    \caption{An overview of the dual process approach, with the data flow and functions summarised for a target task of human segmentation on images. Note: icons are generated using DALLE-3 available via OpenAI ChatGPT under a Plus subscription.}
    \label{fig:method_figure}
\end{figure}


\subsubsection*{The System I module - for adaptable initial solutions and distribution scoring}

Without loss of generality assume that an image sample $x\in\mathcal{X}$ has a human label $y_h\in\mathcal{Y}_h$ (e.g., a binary image mask to delineate an object's boundaries), with $\mathcal{X}$ and $\mathcal{Y}_h$ denoting the image and human label distributions, respectively. 

The \textbf{task-predictor} $f(x; w)=y_t: \mathcal{X} \rightarrow \mathcal{Y}_t$, with network weights $w$, aims to approximate the human label $y_h\in\mathcal{Y}_h$, where the task-predictor's output label prediction belongs to the task-predictor label distribution denoted as $y_t\in\mathcal{Y}_t$.

The \textbf{distribution-discriminator} $d(x, y; \theta)=\hat{z}: \mathcal{X} \times (\mathcal{Y}_h + \mathcal{Y}_t) \rightarrow [0, 1]$, with network weights $\theta$, predicts a score $\hat{z}\in[0, 1]$ between 0 and 1 given an image-label pair $(x, y)$, where $y\in\mathcal{Y}_h + \mathcal{Y}_t$ may be from either the human $\mathcal{Y}_h$ or task-predictor $\mathcal{Y}_t$ label distributions. A score of zero indicates the pair being from the task-predictor label distribution $\mathcal{Y}_t$, and a score of 1 indicates the pair being from the human label distribution $\mathcal{Y}_h$, i.e. 
$z=0$ if $y\in\mathcal{Y}_t$ and $z=1$ if $y\in\mathcal{Y}_h$ where $z \in \{0, 1\}$ is its random variable representing the distinction between human and task-predictor labels.

\textbf{Adversarial training} is used to train the two networks. Let $\mathcal{L}(\cdot): \mathbb{R}^1 \times \mathbb{R}^1 \rightarrow \mathbb{R}^1$ be a function that measures the difference between the distribution-discriminator prediction $\hat{z}$ and the underlying distribution score label $z$. The distribution-discriminator can thus learn to distinguish between task-predictor and human labels. The optimal weights of the distribution-discriminator, $\theta^*$, can be obtained by solving the following optimisation problem, using human labels as well as task-predictor labels:

\begin{align}\label{eq:adv_loss_opt}
    \theta^* = \arg \min_\theta \mathop{\mathbb{E}}_{x\sim\mathcal{X}, ~y\sim\mathcal{Y}_h + \mathcal{Y}_t} ~ [\mathcal{L} (z, ~\hat{z})],\\
    \text{where} ~ \hat{z}=d(x, y; \theta). \nonumber
\end{align}

Note that the loss function for optimising Eq. \ref{eq:adv_loss_opt} that measures the difference between two real numbers is defined as $\mathcal{L}(\cdot): \mathbb{R}^1 \times \mathbb{R}^1 \rightarrow \mathbb{R}^1$, where it takes the form:

\begin{align}
    \mathcal{L}(a, b) = -\left[ a \times \log(b)\right] - \left[ (1-a) \times \log (1-b) \right]
\end{align}

The task-predictor learns to approximate human labels by minimising the difference between the distribution scores of the task-predictor $\hat{z}$ and the distribution scores for human labels (i.e., 1), by updating the task-predictor weights towards the optimal weights $w^*$, in the following optimisation problem:

\begin{align}
w^* = \arg \min_w \mathop{\mathbb{E}}_{x\sim\mathcal{X}} ~ [\mathcal{L} (1, ~\hat{z})],\\
\text{where} ~ \hat{z} = d(x, y_t; \theta), \nonumber\\
\text{and} ~ y_t = f(x; w). \nonumber
\end{align}

In a practical algorithm, the distribution- and task-predictor weight-updating steps are iterated alternatively,
until equilibrium \cite{goodfellow2020generative}.
 
In our work, we assume that during training, the image $x\in\mathcal{X}$ is sampled from a distribution of abdominal CT or MR images, and the human-label $y\in\mathcal{Y}_h$ is a segmentation label, where segmentation tasks include various organ and region-of-interest segmentation tasks. Training across a variety of tasks \cite{nichol2018first} ensures that the task- and distribution-discriminators are generalisable to other tasks on abdominal CT or MR images, during inference. We use the Reptile algorithm \cite{nichol2018first} for meta-learning across various tasks as outlined in Algo\ref{algo:sys_1_train}.

After training the System I module, this can be \textbf{adapted to new tasks} (following Eq. 1 and 2), that did not form part of the training data. This step enables the System I module to output better initial solutions (to be refined by the System II module) and more accurate distribution-discriminator scores, for the new task. This can be achieved using a few human-labelled samples (between 4-16 in our experiments) and following the same update procedure as in training (following Eq. 1 and 2 and Algo. 1, with details on convergence outlined in the Methodological details and definitions section). Then, the adapted task-predictor predicts labels for the new task, for a new sample $x_n$, following $f(x_n; w^*) = y_{t, n}$. The distribution score $\hat{z}\in[0, 1]$ for this particular sample can be computed using the adapted distribution-discriminator $\hat{z} = d(x_n, y_{t, n}; \theta^*)$.

\textbf{Algorithm for training and adapting the System I module:}
We propose to train the System I module across various tasks such that it is adaptable to any new tasks with limited labelled samples. This is achieved by employing the Reptile algorithm during training as summarised in Algo. \ref{algo:sys_1_train}. We can define sample $(x^j, y^j)$ as coming from task $j$, with a total of $J$ tasks available during training.

\begin{algorithm}[!ht]
\caption{Training an adaptable System I module}
\label{algo:sys_1_train}
\SetAlgoLined
\hrulefill\\
    \KwData{$N$ image-label pairs from each of $J$ tasks $\{\{x_n^j, y_{h, n}^j\}_{n=1}^N\}_{j=1}^{J}$}
\KwResult{Trained distribution-discriminator $d(\cdot; \theta^*)$ and task-predictor $f(\cdot; w^*)$}
\hrulefill
\BlankLine
\While{not converged}{
Randomly sample a task $j\in[1, J]$\\
For task $j$, randomly sample $B$ image-label pairs $\{x_n^j, y_{h, n}^j\}_{n=1}^B$\\
\For{$k\leftarrow 1$ \KwTo $K$}{
    Compute task-predictions ${f(x_n; w)}_{n=1}^B = {y_t}_{n=1}^B$\\
\BlankLine
    Combine human and task-predictor labels $\{y\}_{n=1}^{2B} = \{y_{h, n}\}_{n=1}^B \bigcup \{y_{t, n}\}_{n=1}^B$\\     
    Get corresponding underlying distribution scores $\{z\}_{n=1}^{2B} = \{1\}_{n=1}^B \bigcup \{0\}_{n=1}^B$\\
    Assemble image samples $\{x\}_{n=1}^{2B} = \{x\}_{n=1}^{B} \bigcup \{x\}_{n=1}^{B}$\\
\BlankLine
    Compute distribution-predictions ${\hat{z}}_{n=1}^{2B} = {d(x_n, y_n; \theta)}_{n=1}^{2B}$\\
    Take one step of gradient descent to get $\theta_{new}$ following $\arg \min_\theta \mathop{\mathbb{E}}_{~x, ~y} ~ [\mathcal{L} (z, ~\hat{z})]$\\
    Update $\theta \leftarrow \theta + \epsilon (\theta_{new} - \theta)$\\
\BlankLine
     Compute new distribution-predictions ${\hat{z}}_{n=1}^{B} = {d(x_n, y_{t, n}; \theta)}_{n=1}^{B}$\\
     Take one step of gradient descent to get $w_{new}$ following $\arg \min_w \mathop{\mathbb{E}}_{x} ~ [\mathcal{L} (1, ~\hat{z})]$\\
    Update $w \leftarrow w + \epsilon (w_{new} - w)$\\
}
}
\hrulefill
\vspace{0.3cm}
\end{algorithm}

The same scheme as Algo. \ref{algo:sys_1_train} can be used to adapt the System I module to a new task, in which only the new task is sampled, instead of randomly sampling a task from $J$ tasks.

\FloatBarrier

\subsubsection*{The System II module - for an iterative competition between solution refinements}

\textbf{Initial solution using the System I module:} The System II module aims to refine the task-predictor label $y_{t, n}$ for a new sample $x_n$. Since the System II module does not involve any human labels, and works on a per-sample basis, we now omit the subscripts $t$ and $n$ from $y_{t, n}$, in favour of the notationally convenient $y$, to aid readability. So before conducting System II thinking for a new sample $x$, we first compute an initial solution using the System I task-predictor $y = f(x; w^*)$.

\textbf{Competing solution refinement functions:} The auto-competing process in System II thinking involves two competing functions that iteratively refine the task-predictor label. Since the process is iterative, we use a subscript $i$ to denote the iteration number. The competing function is denoted as $h(x, y_{i}; \phi) = y_{i+1}: \mathcal{X} \times \mathcal{Y}_t \rightarrow \mathcal{Y}_t$, with weights $\phi$, where both the pre-refinement label $y_{i}$ and post-refinement label $y_{i+1}$ are assumed to belong to the task-predictor distribution $y_{i}, y_{i+1} \in \mathcal{Y}_t$, even though the actual distribution of refined labels may be different from the task-predictor label distribution. Since there are two competing functions $a$ and $b$ we can denote the two competitors as $h(\cdot; \phi_a)$ and $h(\cdot; \phi_b)$, with the only difference between the two being the function weights. In practice, weights $\phi_b$ are either a copy of $\phi_a$, or $\phi_b$ are sampled from a distribution over all historical weights during training and the weights $\phi_a$ are the most up-to-date weights during training (see further implementation details in `Methodological details and definitions' section).

\textbf{Optimising the competitors during the auto-competing process:} Starting with a new sample $x$ and its task-predictor label $y$ at iteration $i=0$, both competitors refine the label $h(x, y_{i}; \phi_a) = y_{i+1, a}$ and $h(x, y_{i}; \phi_b) = y_{i+1, b}$; here, and in further analyses, subscripts of $a$ or $b$ denote outputs from competitors $a$ or $b$, respectively. The distribution score is measured for both refined labels $d(x, y_{i+1, a}; \theta^*) = \hat{z}_{i, a}$ and $d(x, y_{i+1, b}; \theta^*) = \hat{z}_{i, b}$. These distribution scores are used to formulate a reward for competitor $a$: 

\begin{align}\label{eq:reward}
    R_i = 
    \begin{cases}
    +1 & \textbf{if} ~ \hat{z}_{i, a} \geq \hat{z}_{i, b}\\
    -1 & \text{otherwise}\\
    \end{cases}
\end{align}

This reward is used to train both competitors and update weights $\phi$ towards optimal weights $\phi^*$. For the next iteration $i=i+1$, the refined solution with the higher distribution-discriminator score is set as $y_{i+1}$, meaning that we only use the winning refinement to further refine in the next iterations. We allow iteration of this process until a maximum iteration number $i_{end}$ (see `Methodological details and definitions' for procedure of computing the value of $i_{end}$). We can then reset the refinements and repeat the iterative refinement to collect multiple sets of $\left[{R_i}\right]_{i=0}^{i_{end}}$. The optimisation of weights $\phi$, using experience from both competing sides, can be formulated as the following optimisation problem over multiple collected sets of rewards:

\begin{align}\label{eq:opt_phi}
    \phi^* = \arg\max_\phi \left[ \mathop{\mathbb{E}}_{y_{i, a} \sim h(\cdot; \phi_a)}
    \left[\sum_{i=0}^{i_{end}} R_i \right] + \mathop{\mathbb{E}}_{y_{i, b} \sim h(\cdot; \phi_b)}
    \left[\sum_{i=0}^{i_{end}} -R_i \right] \right]
\end{align}

Solving the above optimisation problem for a single sample $x$ means that no further improvements in the distribution score are possible and thus the fully-refined sample is considered as solved by the System II thinking module. This reward-based training is known as reinforcement learning self-play \cite{silver2018general, saeed2024competing}, where competitors start out with random behaviours which are then refined towards maximising the reward. Practically it may be beneficial to aggregate updates to the competitor weights across a few different samples (up to 8 tested in our experiments), to ensure stable training \cite{liang2018rllib, nair2015massively, kwok2024efficient}. As opposed to directly optimising the task-predictor using the distribution score for a single new sample e.g., using the adversarial training framework to model System II, our auto-competing process which incorporates exploration by the use of reinforcement learning, ensures that the optimisation does not get stuck in local optima, which is a common problem with other adversarial learning frameworks \cite{goodfellow2020generative, creswell2018generative, pan2019recent, saxena2021generative}.

System II refinement is initialised from the prediction of the System I task-predictor to provide a reasonable starting hypothesis for the iterative optimisation process. However, the refinement objective is defined solely by the discriminator score, and therefore the reasoning process is not constrained to remain close to the initial prediction. Instead, candidate refinements are iteratively explored and selected based on their impact on the discriminator score.

For each refinement step, System II selects the single best hypothesis rather than merging the two proposals. This choice is motivated by firstly the fact that many competing refinements are mutually contradictory (e.g., expanding versus contracting a boundary), where merging would produce semantically inconsistent shapes, and secondly the use of relative, rather than absolute, discriminator-based reward assignment, which provides stability against small fluctuations in absolute discriminator scores.

Although the objective of the System II module is to improve segmentation quality, refinement is structured as pairwise competition rather than direct maximisation of the discriminator score. As the discriminator provides a learned proxy rather than the true task metric at inference, relying on absolute score optimisation can be sensitive to calibration noise. Instead, at each iteration two candidate refinements are evaluated comparatively, and only the higher-scoring proposal is enacted and rewarded. This relative reward formulation produces a stable binary reinforcement signal and reduces sensitivity to small fluctuations in the absolute discriminator outputs. Limiting the process to pairwise competition, rather than multiple simultaneous comparisons, maintains a well-conditioned optimisation signal (less impacted by absolute discriminator scores) by avoiding the need to rank or weight multiple competing refinements. Multiple comparisons would increase sensitivity to noise in the discriminator scores, requiring ranking or weighting, leading to less stable optimisation. While such extensions are possible in principle, pairwise comparison provides a more reliable optimisation strategy under a learned reward function. Although the reward is discretised, the magnitude of improvement is implicitly preserved through selection frequency: refinements that yield larger improvements are more consistently chosen across iterations, allowing their effects to accumulate over time. This common comparison-based RL formulation retains robustness to noise in the discriminator by avoiding reliance on absolute score differences, while still favouring substantially better policies through repeated selection.

\textbf{Transitioning tasks from System II to System I:} After a sample is solved by System II thinking, it can be used to update the System I module using the refined label in-place of the human label (following Eq. 1 and 2) to allow the next sample to get a better initial task-predictor label, to reduce the System II thinking time for the subsequent samples. In practice, similar to aggregating updates to the competitor, updates to the System I module can also be aggregated across a few (8 in our experiments) different samples.

\textbf{Preventing disconnects between the discriminator scores and actual task performance:} Two mechanisms ensure alignment between the discriminator and the true task objective to be maintained in the reasoning process. First, before System II reasoning begins, both the task-predictor and distribution-discriminator are adapted using a small set of human-labelled samples specific to the target domain. This ensures that predictions from both networks are task-relevant, minimising the difference in objectives between the domain-of-interest and cross-task pre-training. Furthermore, use of the meta-learning algorithm, Reptile, during pre-training, equips the System I module with the ability to become task-relevant with minimal labelled data.
Once System II reasoning begins, both networks are held fixed, preventing further drift during the refinement process itself. As neither the task-predictor nor the discriminator is updated during System II reasoning, the iterative refinement process cannot introduce or amplify biases in the learned reward signal.
Second, the reward is defined as the relative improvement in discriminator score between competing refinements, rather than relying on the discriminator’s absolute output. Relative rewards are considerably less sensitive to calibration errors in the discriminator, because what matters is the direction of improvement, not the specific numerical value, which is a common strategy used in RL. This comparison-based formulation mitigates the impact of calibration errors in the discriminator because optimisation depends on the direction of improvement between competing refinements rather than the absolute score values. Consequently, small biases in the discriminator are unlikely to accumulate across the iterative reasoning process. Ablation studies removing the relative reward (auto-competition) are presented in Supplementary Table \ref{tab:ablation_res_tab}, which shows that the relative reward formulation offers improved performance.
Note that we intentionally used a few samples for adaptation (e.g., 8 samples as shown in Figure 5c of the main manuscript and Supplementary Table \ref{tab:disc_correlation}). Using a larger adaptation set may further reduce the likelihood of any potential deviation between the discriminator and true segmentation quality. However, our aim was to evaluate the system under a highly sample-efficient regime and demonstrate that strong performance can be achieved even with minimal adaptation data.

\textbf{Algorithm for the auto-competing process:}
The auto-competing process is initialised by first obtaining an initial solution from the already-trained task-predictor. This is then refined by both competitors in the auto-competing process. The distribution scores are computed for both refined solutions and then the one leading to a higher score wins the iteration and is selected to be used for the next iteration. On the next iteration, the refined solution is then further refined by both competitors and the entire process is repeated. Once a specified number of iterations have elapsed, we can update the competitors using gradient ascent informed by the win or loss of the iteration, formulated into a reward. Resetting the refinements and repeating these updates until convergence, we eventually obtain an optimal refinement strategy that leads to the highest distribution score. Then the sample is considered as solved by the System II thinking. This process is summarised in Algo. \ref{algo:sys_2_train}.

\begin{algorithm}[!ht]
\caption{System II solution refinement through and auto-competing mechanism}
\label{algo:sys_2_train}
\SetAlgoLined
\hrulefill\\
\KwData{New sample image $x$}
\KwResult{Optimal refinement strategy $h(\cdot; \phi^*$)}
\hrulefill
\BlankLine
Set $H=0$\\
\BlankLine
\While{not converged}{
Start at $i=0$\\
Sample an initial solution using the task-predictor $f(x; w^*) = y$\\
\BlankLine
Set competitor $a$ weights to $\phi_{a, H}$: $h(\cdot; \phi_{a, H})$\\
Set competitor $b$ weights to $\phi_{b, H_s}$: $h(\cdot; \phi_{b, H_s})$\\
\BlankLine
Sample initial refinement for competitor $a$: $y_{i=0+1, a} = h(x, y; \phi_{a, H})$\\
Sample initial refinement for competitor $b$: $y_{i=0+1, b} = h(x, y; \phi_{b, H_s})$\\
\BlankLine
Compute competitor $a$ distribution score: $d(x, y_{i=0+1, a}; \theta^*) = \hat{z}_{i=0+1, a}$\\
Compute competitor $b$ distribution score: $d(x, y_{i=0+1, a}; \theta^*) = \hat{z}_{i=0+1, b}$\\
\BlankLine
Compute reward $R_1$ using eq \ref{eq:reward}\\
\BlankLine
\textbf{If:} $\hat{z}_{i=0+1, a} \geq \hat{z}_{i=0+1, b}$, \textbf{Then:} Set $y_{i+1} = y_{i=0+1, a}$ \textbf{Else:} Set $y_{i=0+1} = y_{i=0+1, b}$\\
\BlankLine
\For{$i\leftarrow 1$ \KwTo $i_{end}$}{
Sample initial refinement for competitor $a$: $y_{i+1, a} = h(x, y_{i}; \phi_{a, H})$\\
Sample initial refinement for competitor $b$: $y_{i+1, b} = h(x, y_{i}; \phi_{b, H_s})$\\
\BlankLine
Compute competitor $a$ distribution score: $d(x, y_{i+1, a}; \theta^*) = \hat{z}_{i+1, a}$\\
Compute competitor $b$ distribution score: $d(x, y_{i+1, a}; \theta^*) = \hat{z}_{i+1, b}$\\
\BlankLine
Compute reward $R_{i+1}$ using eq \ref{eq:reward}\\
\BlankLine
\textbf{If:} $\hat{z}_{i+1, a} \geq \hat{z}_{i+1, b}$, \textbf{Then:} Set $y_{i+1} = y_{i=0+1, a}$ \textbf{Else:} Set $y_{i=0+1} = y_{n, i=0+1, b}$
}
Once $R_{i=1:i_{end}}$ collected, update weights $\phi$ using Eq. \ref{eq:opt_phi}\\
Set $H = H + 1$
}
\hrulefill
\vspace{0.3cm}
\end{algorithm}

\FloatBarrier


\newpage

\subsection*{Methodological definitions}

\subsubsection*{Competitor sampling}

As outlined in Algo. \ref{algo:sys_2_train}, the variable $H$ is iterated every-time the competitor weights are updated. The weights of competitor $a$ are set to $\phi_{a, H}$: $h(\cdot; \phi_{a, H})$, meaning that these remain as the most up-to-date weights in the optimisation process. However, the weights of competitor $b$ are set to $\phi_{b, H_s}$: $h(\cdot; \phi_{b, h})$ where in the simplest case $H_s=H$ which is commonly known as vanilla reinforcement learning self-play \cite{saeed2024competing}. In our work, we use fictitious self-play to avoid the optimisation being stuck in solutions that do not improve \cite{saeed2024competing}, which means that $H_s \sim 1/H$, denoting the sampling of $H_s$ from a uniform distribution of all values between $0$ and $H$. Practically this means that competitor $b$ is a historic version of competitor $a$ with weights for competitor $b$ sampled from a uniform distribution of all historic weights of competitor $a$.

\subsubsection*{The definition of `refinement'}

Assuming that a grey-scale image $x\in \mathbb{R}^P$ has $P$ pixels, we can assume that the segmentation for such an image would take the form $y\in \{0, 1\}^P$. The refinement of this segmentation that we have thus far summarised as $y_{i_+1} = h(x, y_{i}; \phi)$ can then be modelled in a number of steps as follows:

\begin{align*}
    \text{The refinement}~ y_{i_+1} = h(x, y_{i}; \phi)~ \text{is broken down as follows}:\\
     \text{A segmentation label is denoted as:} ~y_i \in \{0, 1\}^P\\
     \text{In actuality} ~h(x, y_{i}; \phi) = q \in [0, 1]^P ~\text{, values between 0 and 1};\\
    \text{For}~ q\in[0, 1]^P ~\text{set top $p$ values to 1 and rest to 0, to get:}~ q_{rounded}\in \{0, 1\}^P\\
    \text{Apply refinement as an element-wise logical \textbf{XOR} operation:} ~y_{i+1} = y_i ~\oplus~ q_{rounded} \\
\end{align*}

This refinement is carried out using fixed parameters in our work, which were set based on a grid search across 5 different applications. In our work, $p=0.01$ and $i_{end} = p \times P \times 100$.

Our implementation performs refinement at the pixel level using XOR flipping to allow maximum flexibility in the evolving solution space. This binary formulation keeps the refinement action space tractable during iterative reasoning, as the agent only decides whether pixels should be included or excluded from the segmentation rather than determining continuous probability updates. This reduces optimisation complexity and enables stable reinforcement learning convergence, particularly when compute budgets are fixed. While this operates without explicit shape priors or graph/variational constraints, anatomical plausibility is implicitly enforced via the learned distribution-discriminator, which penalises unlikely or anatomically implausible refinements. This avoids imposing hard-coded continuity assumptions, which may conflict with the irregular and heterogeneous morphologies often observed in real cancer pathology. Although refinements operate on binary masks, uncertainty information can still be inferred from the trajectory of refinements across iterations. Regions that undergo frequent modification during the reasoning process correspond to areas of higher ambiguity in the segmentation.

The refinement process operates on the full segmentation mask and does not assume a single-object setting. When multiple objects are present, the distribution-discriminator evaluates the overall segmentation quality across the entire mask. Consequently, candidate refinements that improve some regions while degrading others are compared based on their net effect on the discriminator score, and the refinement yielding the higher overall score is enacted. Because the discriminator is trained on datasets containing multiple objects during cross-task pre-training and task-specific adaptation, it naturally learns to evaluate aggregate or global segmentation quality, even in multi-object scenarios.

\subsubsection*{The definition of convergence}

The convergence in the context of training the System I module is evaluated with respect to the loss $\mathcal{L}$. Let iterations in the while loop in Algo \ref{algo:sys_1_train} be denoted as $o$. The loss at iteration $o$ can thus be denoted as $L_o$. The convergence criterion is defined based on the change in windowed moving average across iterations:

\begin{align}
    1/10 \left[ \sum_{v=0}^{10-1} L_{o-v} - \sum_{v=0}^{10-1} L_{o-10-v}\right] \leq 1e-5
\end{align}

The System II module is considered as converged when the value of $R_i$ stabilises. We can first define $R_{sum} = \sum_{i=1}^{i_{end}} R_i$. Since we iterate $H$ every time $R_{i=1:i_{end}}$ is collected, we can use $H$ as a subscript denoting the sum of rewards at iteration $H$ as $R_{sum, ~H}$. The convergence criterion is then defined as:

\begin{align}
    1/10 \left[ \sum_{v=0}^{10-1} R_{sum,~ H-v} - \sum_{v=0}^{10-1} R_{sum,~ H-10-v}\right] \leq 1
\end{align}

\FloatBarrier

\newpage

\subsection*{Materials and experimental details}

\subsubsection*{Datasets}

For pre-training the System I module, we used a collection of data sets \cite{saeed2024active}, as outlined in Supplementary Table \ref{tab:data sets}. All included data sets are considered relevant to the abdomino-pelvic region and thus the System I learns to be adaptable to other structures or pathologies in the abdomino-pelvic region. The subsequent adaptation to the target task required 8-16 samples.

For data sets that were used for evaluation, we split the data sets, on subject level, into a ratio of 60:40 for the development and held-out sets, to ensure sufficient statistical power for validation. For training or adaptation, we randomly sampled data from the development sets. All results are reported on the held-out sets. 

\begin{table}[!ht]
\small
\centering
\begin{tabular}{|c c c c c c|}
\hline
Dataset & Modality & Structure & Cohort Country & Samples (Held-Out) & Role\\
\hline
\cite{antonelli_decathalon} & CT & Spleen           & United States     &  41 & Training\\
\cite{antonelli_decathalon} & CT & Liver Vessels    & United States     & 303 & Training\\
\cite{synapse_multi_atlas}  & CT & Gallbladder      & United States     & 30 & Training\\
\cite{synapse_multi_atlas}  & CT & Adrenal Gland    & United States     & 30 & Training\\
\cite{synapse_multi_atlas}  & CT & Major Vessels    & United States     & 30 & Training\\
\cite{synapse_multi_atlas}  & CT & Stomach          & United States     & 30 & Training\\
\cite{rister_ct_org}        & CT & Kidneys          & $>$3 countries    & 119 & Training\\
\cite{rister_ct_org}        & CT & Bladder          & $>$3 countries    & 119 & Training\\
\cite{chaos}                & MR & Liver            & Turkey            & 15 & Training\\
\cite{chaos}                & MR & Kidneys          & Turkey            & 15 & Training\\
\cite{chaos}                & MR & Spleen           & Turkey            & 15 & Training\\
\cite{amos}                 & MR & Bladder          & China             & 40 & Training\\
\cite{amos}                 & MR & Gallbladder      & China             & 40 & Training\\
\cite{amos}                 & MR & Prostate         & China             & 40 & Training\\
\cite{amos}                 & MR & Major Vessels    & China             & 40 & Training\\
\cite{antonelli_decathalon} & MR & Prostate         & Netherlands       & 48 & Training\\
\hline
\cite{promis} & mp-MR & Prostate Tumour & United Kingdom & 520 (208) & Evaluation\\
\cite{antonelli_decathalon} & CT  & Liver Tumour & France & 131 (52) & Evaluation\\
\cite{antonelli_decathalon} & CT & Pancreas Tumour & United States & 282 (112) & Evaluation\\
\cite{antonelli_decathalon} & CT & Colon Tumour & United States & 126 (50) & Evaluation\\
\cite{kits} & CT & Kidney Tumour & United States & 489 (195) & Evaluation\\
\hline
\end{tabular}
\caption{A summary of the data sets used for training and evaluation.}
\label{tab:data sets}
\end{table}

\subsubsection*{Network architectures and hyper-parameters}

The hyper-parameters, including the network architectures used are summarised in Supplementary Table \ref{tab:net_hyp_params}.

\begin{table}[!ht]
\small
\centering
\begin{tabular}{|c|c|}
\hline
\multicolumn{2}{|c|}{Task-Predictor - UNet \cite{ronneberger2015u}}\\
\hline
Input Size & $256 \times 256$ \\
Batch Size $B$ & 64\\
Optimiser & Adam\\
Reptile $\epsilon$ & 0.02\\
\hline
\multicolumn{2}{|c|}{Distribution-discriminator - ResNet38}\\
\hline
Input Size & $256 \times 256$ \\
Batch Size & 64\\
Optimiser & Adam\\
\hline
\multicolumn{2}{|c|}{Competing Refinement Functions - UNet \cite{ronneberger2015u}}\\
\hline
Input Size & $P = 256 \times 256$ \\
Batch Size & 8\\
Optimiser & Adam \\
Proportion of pixels to flip per iteration $p$ & $p=$ 0.01\\
Terminal Iteration $i_{end}$ & $p \times P \times 100$ \\
\hline
\end{tabular}
\caption{A summary of hyper-parameter configurations used in this work.}
\label{tab:net_hyp_params}
\end{table}

\FloatBarrier

\newpage

\subsection*{Supplementary results}

\subsubsection*{Detailed results for computer vision tasks}

\paragraph{Digit segmentation from noisy images} For the MNIST digit classification dataset we added Gaussian noise to the images, and thresholded the original clean images at an intensity of 0.5 to generate segmentation ground-truth labels. The System I module was trained to segment digits from noisy images using  digits 0–5. System II thinking, implemented through our proposed framework, was then applied to segment the remaining digits 6-9 (refer to the open-source code repository for further details). For each of the unseen target digits, we used four samples to adapt the System I module prior to System II thinking.

To evaluate performance, we tested the model on 4000 samples of these digits (i.e., digits 6-9). As shown in Figure 2 in the main manuscript, we observed dramatic performance improvements (p-value$=0.0002$, paired Student's t test at a significance level of $\alpha=0.05$) compared to supervised deep learning~\cite{ronneberger2015u} using the same training and adaptation data as our System II approach. Our proposed System II module achieved a Dice score of 97.3$\pm$2.4, compared with the deep learning models which achieved 69.7$\pm$4.3 and 97.2$\pm$2.6, when using 4 and 6000 samples from the target digits for learning, respectively. The latter is an upper-bound performance for these new tasks.

\paragraph{Object segmentation for new classes} Out of 919 classes in the ImageNet-S large-scale unsupervised semantic segmentation (LUSS) data set \cite{gao2022large}, we reserved 40 randomly sampled classes to evaluate our approach. Ten images were randomly sampled from each of these 40 reserved classes (for the purpose of this study, only those classes are considered if they have at least 10 images available per-class). From the remaining classes, ensuring none of the reserved held-out classes were present, 2000 training images were randomly sampled, to train the System I module across different classes, to predict the segmentation of objects. For each of the 40 reserved classes, the System I module was adapted using 4 samples and evaluation was done on the remaining 6 samples. For these $40 \times 6 = 240$ samples, we generated segmentations using our System II thinking approach. 

 As shown in Figure 2 in the main manuscript, our approach obtained a mean intersection over union (mIoU) of 0.691$\pm$0.047 compared to the previous best-performing method \cite{gao2022towards} that obtained 0.538$\pm$0.039 mIoU when using the same training and adaptation data as our System II approach, or 0.645$\pm$0.046 mIoU when using the pre-training strategy as outlined in \cite{gao2022towards} (p-values = $0.0006$ and $0.003$, respectively).

\subsubsection*{Comparison to existing methods for cancer localisation}

In Supplementary Table \ref{tab:seg_res_tab} and \ref{tab:seg_res_tab_miou}, we present detailed results for cancer detection tasks, comparing our System II approach to not only the previous best-performing specialised deep learning models but also to foundation models and widely-used benchmarking U-Nets. The amount of human-labelled data used for training of each method is investigated. Methods that have been pre-trained refer to those using other tasks for learning as summarised in Supplementary Table \ref{tab:data sets}, with the Reptile algorithm \cite{nichol2018first} being used to pre-train across these different tasks. Our method shows superior performance compared to the previously best-performing methods as well as to foundation models and specialised U-Nets trained with much larger datasets (in the order of hundreds of samples), compared to our method which only uses 8-16 samples from the target task.

\begin{table}[!ht]
\centering
\small
\begin{tabular}{|C{1.8cm} C{1cm} C{1cm} | C{1.8cm} C{1.8cm} C{1.8cm} C{1.8cm} C{1.8cm}|}
\hline
Method &            Pre-Train & Labels        & Prostate      & Liver         & Pancreas      & Colon         & Kidney \\
\hline
System II            & Yes       & 8                 & 62.4$\pm$5.1  & \textbf{85.3$\pm$8.6}  & 60.9$\pm$9.7  & \textbf{65.1$\pm$10.7} & \textbf{75.6$\pm$6.3}\\
System II            & Yes       & 12                & \textbf{63.0$\pm$4.8}  & 83.9$\pm$10.2 & 63.6$\pm$8.9  & 63.3$\pm$10.2 & 74.9$\pm$5.9\\
System II            & Yes       & 16                & 62.8$\pm$5.8  & 84.3$\pm$9.1  & \textbf{66.2$\pm$9.4}  & 64.7$\pm$11.0 & 74.1$\pm$6.2\\
\hline
UNet \cite{ronneberger2015u}               & Yes       & 8                  & 24.9$\pm$5.5  & 32.7$\pm$10.1 & 24.6$\pm$8.5  & 34.8$\pm$10.7  & 31.6$\pm$6.2\\
UNet \cite{ronneberger2015u}               & Yes       & 12                 & 28.5$\pm$5.3  & 40.2$\pm$11.3 & 31.2$\pm$9.2  & 32.4$\pm$11.6  & 34.1$\pm$5.3\\
UNet \cite{ronneberger2015u}               & Yes       & 16                 & 29.8$\pm$4.7  & 43.1$\pm$9.4  & 45.0$\pm$8.7  & 41.7$\pm$11.1  & 35.3$\pm$5.7\\
UNet \cite{ronneberger2015u}               & Yes       & $>$100             & 41.7$\pm$5.9  & 74.2$\pm$9.7  & 56.1$\pm$9.1  & 53.9$\pm$11.9  & 71.8$\pm$6.8\\
SUNeTR \cite{hatamizadeh2021swin}               & Yes       & 8                  & 25.6$\pm$5.3  & 35.8$\pm$10.6 & 26.3$\pm$8.7  & 34.9$\pm$9.8  & 34.1$\pm$6.6\\
SUNeTR \cite{hatamizadeh2021swin}               & Yes       & 12                 & 28.7$\pm$5.9  & 39.3$\pm$10.4 & 33.6$\pm$9.8  & 36.4$\pm$11.7  & 35.1$\pm$6.5\\
SUNeTR \cite{hatamizadeh2021swin}               & Yes       & 16                 & 34.5$\pm$4.9  & 45.9$\pm$9.9  & 46.2$\pm$9.1  & 45.3$\pm$10.6  & 43.4$\pm$5.8\\
SUNeTR \cite{hatamizadeh2021swin}               & Yes       & $>$100             & 41.9$\pm$5.4  & 76.7$\pm$10.2  & 57.3$\pm$8.9  & 52.8$\pm$11.2  & 70.8$\pm$5.9\\
SegNeXt \cite{guo2022segnext}               & Yes       & 8                  & 23.5$\pm$4.5  & 31.0$\pm$11.8 & 25.9$\pm$9.8  & 32.5$\pm$11.9  & 31.2$\pm$6.7\\
SegNeXt \cite{guo2022segnext}               & Yes       & 12                 & 29.6$\pm$5.2  & 41.4$\pm$9.7 & 32.8$\pm$8.4  & 38.3$\pm$11.2  & 35.1$\pm$6.3\\
SegNeXt \cite{guo2022segnext}               & Yes       & 16                 & 33.3$\pm$5.8  & 44.7$\pm$10.2 & 48.4$\pm$9.3  & 42.9$\pm$11.4  & 47.8$\pm$6.1\\
SegNeXt \cite{guo2022segnext}               & Yes       & $>$100             & 40.2$\pm$4.9  & 74.0$\pm$10.5  & 58.2$\pm$8.9  & 56.7$\pm$10.7  & 70.9$\pm$5.8\\
\hline
UNet \cite{ronneberger2015u}               & No        & 8                  & 18.3$\pm$5.7  & 20.8$\pm$8.9  & 20.5$\pm$9.6  & 28.3$\pm$11.8  & 30.8$\pm$6.6\\
UNet \cite{ronneberger2015u}               & No        & 12                 & 22.7$\pm$4.6  & 27.4$\pm$9.7  & 33.9$\pm$9.3  & 30.1$\pm$11.5  & 32.4$\pm$5.2\\
UNet \cite{ronneberger2015u}               & No        & 16                 & 26.8$\pm$4.9  & 34.7$\pm$10.5 & 41.4$\pm$8.5  & 40.9$\pm$10.3  & 39.3$\pm$6.3\\
UNet \cite{ronneberger2015u}               & No        & $>$100             & 37.6$\pm$4.9  & 74.5$\pm$10.4 & 54.0$\pm$9.9  & 55.2$\pm$10.2  & 70.2$\pm$5.6\\
\hline
Dv2 \cite{oquab2023dinov2}        & Yes       & 8         & 38.4$\pm$4.9  & 37.4$\pm$9.1  & 37.0$\pm$8.6  & 36.1$\pm$10.4  & 38.1$\pm$6.2\\
Dv2 \cite{oquab2023dinov2}        & Yes       & 12        & 37.9$\pm$4.3  & 39.3$\pm$8.6  & 37.2$\pm$8.9  & 37.0$\pm$10.5  & 37.8$\pm$5.1\\
Dv2 \cite{oquab2023dinov2}        & Yes       & 16        & 38.5$\pm$5.6  & 41.6$\pm$9.7  & 38.7$\pm$9.3  & 39.2$\pm$11.5  & 40.9$\pm$5.7\\
Dv2 \cite{oquab2023dinov2}        & Yes       & $>$100    & 40.8$\pm$5.6  & 73.5$\pm$10.3 & 55.9$\pm$9.1  & 53.4$\pm$10.3  & 72.5$\pm$6.8\\
USeg \cite{butoi2023universeg}    & Yes       & 8         & 27.6$\pm$5.3  & -             & 33.2$\pm$8.2  & 39.1$\pm$11.9  & -\\
USeg \cite{butoi2023universeg}    & Yes       & 12        & 34.7$\pm$5.8  & -             & 34.7$\pm$8.9  & 40.5$\pm$11.3  & -\\
USeg \cite{butoi2023universeg}    & Yes       & 16        & 36.0$\pm$4.7  & -             & 40.6$\pm$8.8  & 42.9$\pm$10.8 & -\\
USeg \cite{butoi2023universeg}    & Yes       & $>$100    & 42.1$\pm$5.1  & -             & 60.7$\pm$9.5  & 61.5$\pm$10.1  & -\\
\hline
nnUNet \cite{isensee2021nnu}              & No        & $>$100            & 42.3$\pm$5.6  & -             & -             & -             & -             \\
CLIP  \cite{liu2023clip}               & Yes       & $>$100            & -             & 79.4$\pm$8.1 & 62.3$\pm$9.8  & 63.1$\pm$10.6  & -             \\
ASeg  \cite{myronenko2023automated}           & No        & $>$100            & -             & -             & -             & -             & \textbf{76.4$\pm$5.5}\\
\hline
\end{tabular}
\caption{Cancer lesion segmentation results (Dice score, as widely reported in medical imaging literature). Where we specify $>100$ human labels, the full train sets from the respective tasks have been used. \textbf{First block:} results from our System II approach; \textbf{Second block:} UNet model \cite{ronneberger2015u} pre-trained on the same data as our System I module using the Reptile algorithm and fine-tuned using human labels for each application; \textbf{Third block:} UNet model without pre-training; \textbf{Fourth block:} foundation models (Dv2 is DINO-v2 \cite{oquab2023dinov2} encoder fine-tuned on 50\% of training data and segmentation head fine-tuned on remaining data; with labelled samples for each application used to fine-tune the entire model before inference; USeg is the UniverSeg \cite{butoi2023universeg} model either fine-tuned or using support samples for each application - applications used in training of USeg were not considered); \textbf{Fifth block:} previous best-performing methods for each application. Where results are emboldened in more than one block, this indicates that statistical significance was not found for the comparison. Pre-train indicates the use of pre-training sets before training for the target task.}
\label{tab:seg_res_tab}
\end{table}


\begin{table}[!ht]
\centering
\small
\begin{tabular}{|C{1.8cm} C{1cm} C{1cm} | C{1.8cm} C{1.8cm} C{1.8cm} C{1.8cm} C{1.8cm}|}
\hline
Method &            Pre-Train & Labels        & Prostate      & Liver         & Pancreas      & Colon         & Kidney \\
\hline
System II            & Yes       & 8          & 45.3$\pm$5.4  & \textbf{74.4$\pm$13.1} & 43.8$\pm$10.0 & \textbf{48.3$\pm$11.8} & \textbf{60.8$\pm$8.1}\\
System II            & Yes       & 12         & \textbf{46.0$\pm$5.1} & 72.3$\pm$15.1 & 46.6$\pm$9.6  & 46.3$\pm$10.9 & 59.9$\pm$7.5\\
System II            & Yes       & 16         & 45.8$\pm$6.2  & 72.9$\pm$13.6 & \textbf{49.5$\pm$10.5} & 47.8$\pm$12.0 & 58.9$\pm$7.8\\
\hline
UNet \cite{ronneberger2015u}               & Yes       & 8          & 14.2$\pm$3.6  & 19.5$\pm$7.2  & 14.0$\pm$5.5  & 21.1$\pm$7.8  & 18.8$\pm$4.4\\
UNet \cite{ronneberger2015u}               & Yes       & 12         & 16.6$\pm$3.6  & 25.2$\pm$8.9  & 18.5$\pm$6.5  & 19.3$\pm$8.3  & 20.6$\pm$3.9\\
UNet \cite{ronneberger2015u}               & Yes       & 16         & 17.5$\pm$3.2  & 27.5$\pm$7.6  & 29.0$\pm$7.2  & 26.3$\pm$8.9  & 21.4$\pm$4.2\\
UNet \cite{ronneberger2015u}               & Yes       & $>$100     & 26.3$\pm$4.7  & 59.0$\pm$12.3 & 39.0$\pm$8.8  & 36.9$\pm$11.2 & 56.0$\pm$8.3\\
SUNeTR \cite{hatamizadeh2021swin}       & Yes       & 8          & 14.7$\pm$3.5  & 21.8$\pm$7.9  & 15.1$\pm$5.8  & 21.1$\pm$7.2  & 20.6$\pm$4.8\\
SUNeTR \cite{hatamizadeh2021swin}       & Yes       & 12         & 16.8$\pm$4.0  & 24.5$\pm$8.1  & 20.2$\pm$7.1  & 22.2$\pm$8.7  & 21.3$\pm$4.8\\
SUNeTR \cite{hatamizadeh2021swin}       & Yes       & 16         & 20.8$\pm$3.6  & 29.8$\pm$8.3  & 30.0$\pm$7.7  & 29.3$\pm$8.9  & 27.7$\pm$4.7\\
SUNeTR \cite{hatamizadeh2021swin}       & Yes       & $>$100     & 26.5$\pm$4.3  & 62.2$\pm$13.4 & 40.2$\pm$8.7  & 35.9$\pm$10.3 & 54.8$\pm$7.1\\
SegNeXt \cite{guo2022segnext}               & Yes       & 8         & 13.3$\pm$2.9  & 18.3$\pm$8.3  & 14.9$\pm$6.5  & 19.4$\pm$8.5  & 18.5$\pm$4.7\\
SegNeXt \cite{guo2022segnext}               & Yes       & 12        & 17.4$\pm$3.6  & 26.1$\pm$7.7  & 19.6$\pm$6.0  & 23.7$\pm$8.6  & 21.3$\pm$4.6\\
SegNeXt \cite{guo2022segnext}               & Yes       & 16        & 20.0$\pm$4.2  & 28.8$\pm$8.5  & 31.9$\pm$8.1  & 27.3$\pm$9.2  & 31.4$\pm$5.3\\
SegNeXt \cite{guo2022segnext}               & Yes       & $>$100    & 25.2$\pm$3.8  & 58.7$\pm$13.2 & 41.0$\pm$8.9  & 39.6$\pm$10.4 & 54.9$\pm$7.0\\
\hline
UNet \cite{ronneberger2015u}               & No        & 8           & 10.1$\pm$3.5  & 11.6$\pm$5.5  & 11.4$\pm$6.0  & 16.5$\pm$8.0  & 18.2$\pm$4.6\\
UNet \cite{ronneberger2015u}               & No        & 12          & 12.8$\pm$2.9  & 15.9$\pm$6.5  & 20.4$\pm$6.7  & 17.7$\pm$8.0  & 19.3$\pm$3.7\\
UNet \cite{ronneberger2015u}               & No        & 16          & 15.5$\pm$3.3  & 21.0$\pm$7.7  & 26.1$\pm$6.8  & 25.7$\pm$8.1  & 24.5$\pm$4.9\\
UNet \cite{ronneberger2015u}               & No        & $>$100      & 23.2$\pm$3.7  & 59.4$\pm$13.2 & 37.0$\pm$9.3  & 38.1$\pm$9.7  & 54.1$\pm$6.6\\
\hline
Dv2 \cite{oquab2023dinov2}        & Yes       & 8         & 23.8$\pm$3.8  & 23.0$\pm$6.9  & 22.7$\pm$6.5  & 22.0$\pm$7.7  & 23.5$\pm$4.7\\
Dv2 \cite{oquab2023dinov2}        & Yes       & 12        & 23.4$\pm$3.3  & 24.5$\pm$6.7  & 22.9$\pm$6.7  & 22.7$\pm$7.9  & 23.3$\pm$3.9\\
Dv2 \cite{oquab2023dinov2}        & Yes       & 16        & 23.8$\pm$4.3  & 26.3$\pm$7.7  & 24.0$\pm$7.1  & 24.4$\pm$8.9  & 25.7$\pm$4.5\\
Dv2 \cite{oquab2023dinov2}        & Yes       & $>$100    & 25.6$\pm$4.4  & 58.1$\pm$12.9 & 38.8$\pm$8.8  & 36.4$\pm$9.6  & 56.9$\pm$8.4\\
USeg \cite{butoi2023universeg}    & Yes       & 8         & 16.0$\pm$3.6  & -             & 19.9$\pm$5.9  & 24.3$\pm$9.2  & -\\
USeg \cite{butoi2023universeg}    & Yes       & 12        & 21.0$\pm$4.2  & -             & 21.0$\pm$6.5  & 25.4$\pm$8.9  & -\\
USeg \cite{butoi2023universeg}    & Yes       & 16        & 22.0$\pm$3.5  & -             & 25.5$\pm$6.9  & 27.3$\pm$8.8  & -\\
USeg \cite{butoi2023universeg}    & Yes       & $>$100    & 26.7$\pm$4.1  & -             & 43.6$\pm$9.8  & 44.4$\pm$10.5 & -\\
\hline
nnUNet \cite{isensee2021nnu}              & No        & $>$100       & 26.8$\pm$4.5  & -             & -             & -             & -             \\
CLIP  \cite{liu2023clip}               & Yes       & $>$100            & -             & 65.8$\pm$11.1 & 45.2$\pm$10.3 & 46.1$\pm$11.3 & -             \\
ASeg  \cite{myronenko2023automated}           & No        & $>$100            & -             & -             & -             & -             & \textbf{61.8$\pm$7.2}\\
\hline
\end{tabular}
\caption{The mean intersection over union (mIoU) metric reported for cancer lesion segmentation, corresponding to the above Supplementary Table \ref{tab:seg_res_tab} (where Dice was reported).}
\label{tab:seg_res_tab_miou}
\end{table}

\FloatBarrier

\subsubsection*{Retrospective evaluation against human experts for prostate cancer detection}

We evaluate performance against uro-radiology-specialised radiologists in prostate cancer detection, in which ground-truth labels are available from independent histpathology examinations. We trained our System II cognition using 16 samples with histopathology labels derived from prostate template-mapping biopsy to indicate cancer presence in one of the twenty Barzell zones within the gland~\cite{ahmed2017diagnostic}. Each zone had a label of one to indicate positive cancer with a Gleason grade $\geq$ 3+4, and zero otherwise. We observed an across-zone sensitivity and specificity of 0.774 and 0.582. From the zonal predictions, we predicted patient-wise classification (by predicting cancer presence when any single zone was deemed cancer-positive). We used zonal prediction to compute patient-wise classification, as zonal prediction can be formulated as low-resolution segmentation and thus does not require retraining of the System I module for classification tasks, in addition to its potential spatial interpretability.

\subsubsection*{Ablation and robustness studies}

For the task of liver tumour segmentation, we also present ablation studies in Supplementary Table \ref{tab:ablation_res_tab}, where we remove components including adversarial training and RL from our framework to evaluate their impact on task performance. When adversarial training is omitted, the two networks in the System I module are trained separately, where the task-predictor is first trained to perform the task and then a distribution-discriminator is separately trained after training the task-predictor. When auto-competition is omitted, this refers to using a single agent RL framework with a reward based on distribution-discriminator score, which is directly maximised, as opposed to difference in scores between the predictions from two competing agents. Where RL is omitted, the task-predictor is optimised directly per-sample by maximising the distribution-discriminator score as in the adversarial learning for each new sample. The results show that all of these components contribute positively towards performance. 

In particular, it is interesting to note that the removal of adversarial training led to a marked reduction in performance, which highlights the interdependent nature of the solution prediction and distribution discrimination. During adversarial training, the discriminator continuously adapts to the evolving predictor outputs, while the predictor updates to better match the learned solution distribution. This coupling enables System I to capture the interdependence between solution quality and distributional plausibility, which is lost when the components are trained separately as conventional neural networks.

\begin{table}[!ht]
\small
\centering
\begin{tabular}{|C{2.0cm} C{2.0cm} C{2.0cm} | C{2.0cm} C{2.0cm} | C{2.5cm}|}
\hline
\multicolumn{3}{|c|}{System I module} & \multicolumn{2}{c|}{System II module} &\\
\hline
Adversarial Training   & Reptile    & Pre-Training  & Auto-competition  & RL        & Performance \\
\hline
\cmark                 & \cmark     & \cmark        & \cmark             & \cmark   & \textbf{85.3$\pm$4.3} \\
\hline
\cmark                 & \xmark     & \xmark        & \cmark             & \cmark   & 57.9$\pm$5.1 \\
\xmark                 & \cmark     & \cmark        & \cmark             & \cmark   & 75.7$\pm$5.8 \\
\cmark                 & \xmark     & \cmark        & \cmark             & \cmark   & \textbf{83.7$\pm$4.7} \\
\xmark                 & \xmark     & \xmark        & \cmark             & \cmark   & 53.1$\pm$4.5 \\
\hline
\cmark                 & \cmark     & \cmark        & \xmark             & \cmark   & 78.5$\pm$6.3 \\
\cmark                 & \cmark     & \cmark        & \xmark             & \xmark   & 64.6$\pm$4.6 \\
\hline
\cmark                 & \xmark     & \xmark        & \xmark             & \xmark   & 38.7$\pm$4.9 \\
\hline
\end{tabular}
\caption{Ablation study - removing components from the framework to test efficacy of different approaches. The standard deviation is reported to measure variance across 4 runs with different seeds.}
\label{tab:ablation_res_tab}
\end{table}

In addition to these ablations, we also study the impact of different hyper-parameters and amounts of labelled data, as summarised below.

\textbf{Amounts of human-labelled data for adaptation:}
To analyse the alignment between task performance and the discriminator score, we investigate the correlation between Dice score and the discriminator score, for the task of prostate cancer segmentation, when using different amounts of human labelled data for adaptation of the System I module. The results are presented in Supplementary Table \ref{tab:disc_correlation} below, where we measure the Pearson's correlation coefficient (and standard error) between the Dice score (measured between task-predictor labels, from a task-predictor adapted with 32 samples, and human labels) and discriminator score, across 100 human-labelled samples in the holdout set. We observe that approximately 8 samples allow for correlations approaching the level of alignment observed for much larger sets of labelled data. Similar behaviour was observed across all evaluated tasks. When using eight adaptation samples, the Pearson correlation between discriminator score and Dice performance was $>$0.78 for all evaluated tasks, indicating that the learned reward signal remained consistently aligned with the true task objective across domains.

\begin{table}[!ht]
\small
\centering
\begin{tabular}{|c|c|}
\hline
Human labels & Correlation \\
\hline
0 & 0.21 $\pm$ 0.09\\
4 & 0.53  $\pm$ 0.09\\
8 &  0.81 $\pm$ 0.06\\
12  & 0.83 $\pm$ 0.06\\
16  & 0.79 $\pm$ 0.06\\
20  & 0.84 $\pm$ 0.05\\
24  & 0.83 $\pm$ 0.06\\
28  & 0.81 $\pm$ 0.06\\
32  & 0.80 $\pm$ 0.06\\
\hline
\end{tabular}
\caption{Impact of human-labelled data for adaptation of the System I module on the alignment between the distribution discriminator scores and actual performance (Dice).}
\label{tab:disc_correlation}
\end{table}

\noindent\textbf{Impact of annotation noise:} We study the impact of noisy human annotated data on the performance of our proposed method compared to deep learning. The detailed quantitative results are presented in Supplementary Table \ref{tab:noisy_data}. Note that we added random deformations to the segmentation labels to simulate human error in annotation, the biggest source of noise in our data. The segmentation masks were deformed using a grid of 3$\times$3 control points, each with a random displacement of 0-40 pixels or 0-20mm, with dense displacements computed using linear interpolation. The images themselves were kept unchanged, as they already contain real-world noise sources (e.g., motion artefacts, probe variability and sensor noise), and because realistic simulation of imaging noise remains an open research problem without standardised models. Additionally, For different proportions of noisy adaptation samples, we compared the proposed method with a deep-learning-based segmentation baseline. As shown below, both methods experience performance degradation as noise increases, but our approach shows a more gradual decline, indicating better robustness to annotation outliers. 


\begin{table}[!ht]
\small
\centering
\begin{tabular}{|c|c|c|}
\hline
Percentage corrupted & System II & DL - Pre-trained \\
\hline
0 & 62.7$\pm$5.7 & 40.6$\pm$4.9\\
25 & 60.1$\pm$5.2 & 31.4$\pm$5.9\\
50 & 51.3$\pm$5.9 & 26.8$\pm$5.6\\
75 & 40.2$\pm$5.4 & 22.6$\pm$4.8\\
100 & 30.2$\pm$5.2 & 21.3$\pm$5.5\\
\hline
\end{tabular}
\caption{Studying the impact of noisy data on performance for the prostate cancer segmentation task. The `percentage corrupted' refers to the percentage of adaptation data to which noise was added out of a total of 32 adaptation samples.}
\label{tab:noisy_data}
\end{table}

\begin{figure}[!ht]
    \centering
    \includegraphics[width=0.5\linewidth]{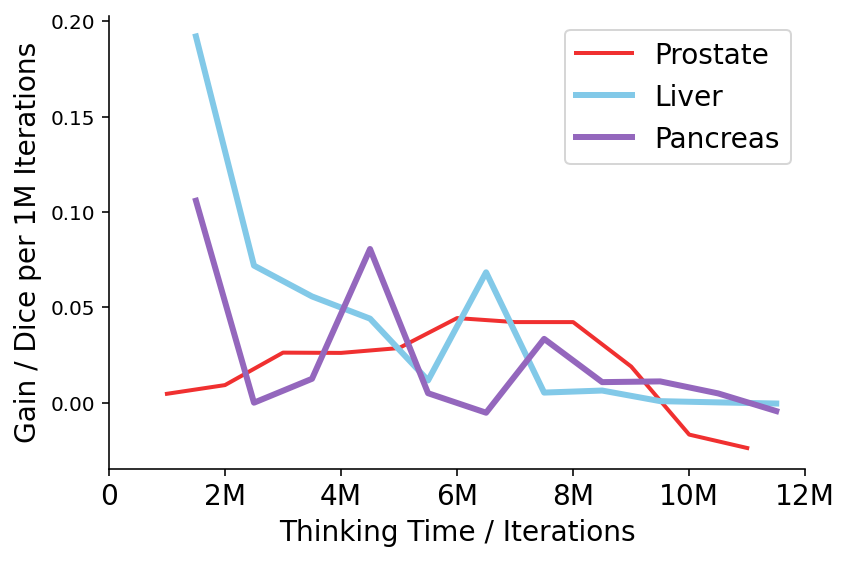}
    \caption{Performance gain per-sample against number of compute iterations}
    \label{fig:gain}
\end{figure}

\noindent\textbf{Impact of compute on per-sample performance:} Supplementary Figure \ref{fig:gain} shows the performance gain per-sample against the number of iterations. We observe an overall reduction in gain, towards zero, with an increasing number of iterations. This highlights that the proposed auto-competing mechanism converges to stable values as the refinement in the System II module progresses.

\noindent\textbf{Analysis of hyper-parameters:} The analysis of hyper-parameters $p$ (the proportion of pixels to flip per iteration) and $i_{\text{end}}$ is presented in Supplementary Table \ref{tab:hyper-param}. Note that we observed a high dependence of the optimal values for both hyper-parameters on the input size or number of pixels in the image $P$. As such, we defined $i_{\text{end}}$ as a function of $P$. As the input size stays consistent across datasets in our work, we observed a consistent optimal value for both $p$ and $i_{\text{end}}$ across the evaluation datasets. However, for other datasets, in future work, these parameters may need to be tuned according to the input or ROI size.

\begin{table}[!ht]
\small
\centering
\begin{tabular}{|c|c|c|c|c|}
\hline
Proportion $p$ & Termination $i_{\text{end}}$ & Prostate & Liver & Pancreas\\
\hline
0.001 & $p\times P\times 1$     & 56.8$\pm$5.7 & 74.3$\pm$8.8 & -\\
0.005 & $p\times P\times 1$     & 60.4$\pm$4.9 & 79.1$\pm$9.2 & 62.9$\pm$8.8\\
0.010 & $p\times P\times 1$     & 61.7$\pm$5.8 & 81.5$\pm$9.7 & 65.2$\pm$9.2\\
0.050 & $p\times P\times 1$     & 61.3$\pm$6.3 & 78.7$\pm$9.1 & 62.4$\pm$9.9\\
0.100 & $p\times P\times 1$     & 50.4$\pm$7.8 & 65.3$\pm$11.7 & -\\
\hline
0.010 & $p\times P\times 1$     & 61.7$\pm$5.8 & 81.5$\pm$9.7 & -\\
0.010 & $p\times P\times 10$    & 60.6$\pm$5.4 & 83.7$\pm$9.5 & 65.6$\pm$9.1\\
0.010 & $p\times P\times 100$   & 63.0$\pm$4.8 & 85.3$\pm$8.6 & 66.2$\pm$9.4 \\
0.010 & $p\times P\times 1000$  & 58.4$\pm$5.9 & 80.5$\pm$9.5 & 61.1$\pm$8.7\\
0.010 & $p\times P\times 10000$ & 54.1$\pm$6.7 & 78.4$\pm$9.4 & -\\
\hline
\end{tabular}
\caption{Impact of hyper-parameters on performance for models adapted using 8 labelled samples.}
\label{tab:hyper-param}
\end{table}

\subsubsection*{Uncertainty quantification with respect to data}

To quantify the uncertainty in the computed segmentation using our proposed approach, we reserved a set of 32 samples from the liver tumour segmentation dataset. To each of these samples and their ground-truth human expert labels, we applied 32 random affine transformations. The transformations altered translation, scale, shear, rotation or a combination of these corresponding to application-specific transformations that may be observed due to shifts in imaging camera positions, variations in patient positions and variations in anatomical appearances due to noise from processes such as breathing. The standard deviation in performance (Dice score) across these affine transforms is computed per-sample and then averaged across the 32 reserved samples to obtain an estimate of the aleatoric uncertainty with respect to variations in input data.

\begin{figure}[!ht]
    \centering
    \includegraphics[width=0.6\linewidth]{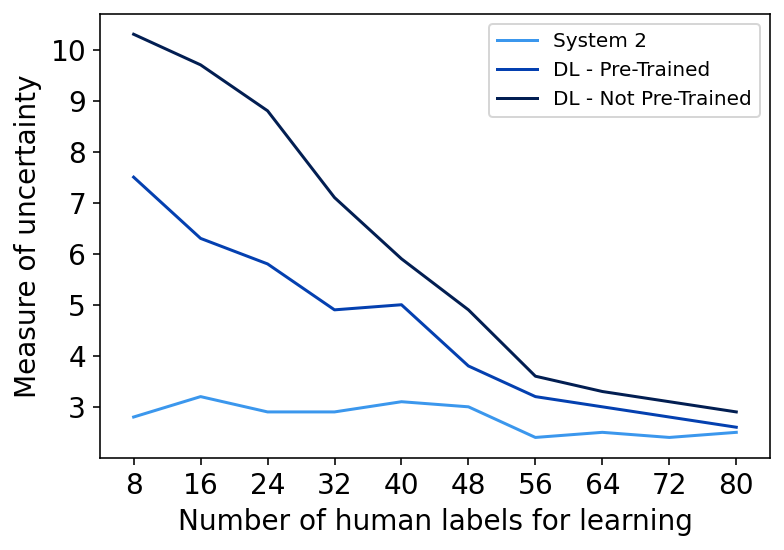}
    \caption{Visualising uncertainty against number of human labelled samples used for learning, from the proposed and compared methods, in the liver tumour segmentation task.}
    \label{fig:uncert}
\end{figure}

Supplementary Figure \ref{fig:uncert} present plots of these measures of uncertainty for our proposed System II machine cognition as well as other commonly used methods for segmentation. All methods show a trend of decreasing uncertainty with an increase in the number of human labelled samples used, however, our proposed System II method shows the most stable measure of uncertainty which demonstrates its reduced reliance on human labels compared to other methods.

\subsubsection*{Computational cost}

We summarise the carbon footprint of training and evaluating our proposed System II machine cognition framework in Supplementary Table \ref{tab:carbon}. For computing the total power consumption or kg of CO2 produced, the given measures can be multiplied by the GPU hours.

\begin{table}[!ht]
\centering
\small
\begin{tabular}{|c |c |c |c c c c c c |c |c |}
\hline
Task & \multicolumn{8}{|c|}{GPU Hours} & Power & Carbon\\ 
\cline{2-9}
 & Pre-Train    & Adapt & \multicolumn{6}{c|}{Inference / Reasoning Time} &  (W) & (kg CO2 / \\ 
\cline{4-9}
 &   &  & \multicolumn{6}{c|}{Experience (Number of Samples)} &   &  GPU hour)\\ 
 &              &       & 0 & 8 & 16 & 24 & 32 & 40 &  &  \\ 
\hline
Prostate& 1152 & 16     & 41 & 36 & 24 & 16 & 6 & 6 & 300 W & 0.13 \\
Liver   & 1152 & 18     & 37 & 34 & 20 & 8 & 8 & 7 & 300 W & 0.13 \\
Pancreas& 1152 & 13     & 42 & 39 & 19 & 9 & 7 & 8 & 300 W & 0.13 \\
Colon   & 1152 & 15     & 33 & 30 & 11 & 3 & 4 & 3 & 300 W & 0.13 \\
Kidney  & 1152 & 21     & 29 & 23 & 17 & 10 & 4 & 2 & 300 W & 0.13 \\
\hline
\end{tabular}
\caption{Carbon footprints estimated on Nvidia V100 GPUs.}
\label{tab:carbon}
\end{table}

\subsubsection*{Adapting the method for classification or regression}

Our System II method can readily be adapted to other machine learning tasks by modifying the definition of the solution space and its refinement. 

\paragraph{Classification:}

For classification, assume that labels are standardised to the continuous range $y\in\{0, 1\}$, representing class probabilities, with 1 being a 100\% probability of belonging to the class-of-interest. For multi-class classification, we can reformulate the problem to multiple binary classification problems (classifying whether a sample belongs to class-of-interest or not) similar to \cite{saeed2024competing}. To enable continuous values as labels, we apply mixup \cite{liang2018understanding, zhang2017mixup} to training images and their labels, mixing samples from the class-of-interest with out-of-class samples. The formula for mixing is $(\lambda) z_i + (1-\lambda)z_j$, where $z_i$ and $z_j$ are the class-of-interest sample and out-of-class sample respectively, and $\lambda$ is a mixing factor randomly sampled between 0-1. The mixing is applied to random class-of-interest and out-of-class samples to generate continuous labels for training. Once the System I module is trained with this mixed-up data, the task-predictor learns to output continuous values $f(x; w) = y\in\{0, 1\}$ for each image sample. 

During System II thinking, the competing function $h(x, y_i; \phi)$ is modified to output values in the range $\{-0.01, +0.01\}$ (instead of the flipping procedure described in the definition of refinement), which can then be added to the label $y_i$ in order to obtain $y_{i+1}$. Note that $y_0$ is initialised as 0.5, which represents a 50\% probability of the sample belonging to the class-of-interest. Then during System II thinking the competing functions compete to refine the class probability for the sample, which is especially useful for ambiguous samples, to assign a representative class probability.

\paragraph{MNIST digit classification:}

We utilised the same training data as the MNIST segmentation problem discussed in the preliminary results section. This data consisted of 4000 samples from each of digits 0-9, with artificial Gaussian noise added to the images. Data from digits 0-5 was used for training the System I module for digit classification. From digits 6-9, 4 samples each were used to adapt the task predictor and distribution-discriminator prior to System II thinking. For each digit, during training and inference, due to the task binarisation process, the task was split into correct digit vs incorrect digit as opposed to a direct multi-class classification (i.e., the tasks were digit 6 vs not 6, digit 7 vs not 7, digit 8 vs not 8, and digit 9 vs not 9, instead of the multi-class representation digit 6 vs 7 vs 8 vs 9). Performing System II refinement, we observed an accuracy of 0.953, compared to 0.829 from the alternative supervised deep learning models using the same training data. The results are summarised in Supplementary Table \ref{tab:class_reg}.

\paragraph{IQ test puzzle classification}

This task is used to demonstrate spatial reasoning. We used the MMIQ \cite{cai2025mm} dataset for this experiment. The task predictor is configured to intake a 5-channel image with the first channel being the input puzzle (with a missing square) and the other channels being possible solution options A, B, C or D inserted into the missing square to build a completed puzzle. The System I module is trained with IQ puzzles spanning geometric transformations, spatial relationships, and logical patterns with 395, 102 and 335 puzzles in each category. The multi-class classification problem is therefore to pick the correct solution out of the 4 presented options, where this multi-class task is binarised in the same way as for the MNIST experiment.

Inference is performed on unseen movement puzzles, examples requiring pattern continuation in space. The System I module is adapted using 4 samples from these movement puzzles prior to System II applied to 379 puzzles, across which results are reported. The results, exceeding performance compared to state-of-the-art systems such as ChatGPT are presented in Supplementary Table \ref{tab:class_reg} and they highlight that language based reasoning developed in multimodal systems does not readily extend to visual tasks.

\paragraph{Regression:}

Regression follows a similar formulation whereby labels are standardised to the continuous range $y\in\{0, 1\}^N$, representing target values for a multiple regression problem. The System I module can be trained with these target values and learns to predict $f(x; w) = y\in\{0, 1\}^N$.

Similar to classification, during System II thinking, the competing function $h(x, y_i; \phi)$ is modified to output values in the range $\{-0.01, +0.01\}$, which can be added to the label $y_i$ in order to obtain $y_{i+1}$. Similarly, $y_0$ is initialised as 0.5. Then during System II thinking the competing functions compete to refine the target values, to reflect correct targets.

\paragraph{Depth estimation regression}

As an example of scene understanding, we evaluate the framework for depth estimation. We used the NYU Depth V2 dataset for depth estimation in natural RGB images \cite{silberman2012indoor}. The target values are the pixel-wise depths for the images and thus the target shape corresponds to the image shape $N=640\times480$. The depths were normalised to range 0-1. 

The System I module was trained with data from 20 scenes (e.g., bedroom, kitchen, bookstore, basement etc.). It was adapted to the scene `bathroom' using 4 samples. System II refinement was performed for 64 samples, on which the results are reported in Supplementary Table \ref{tab:class_reg}. We demonstrate superior performance to conventional approaches given the same amount of training data.

\begin{table}[!ht]
\centering
\small
\begin{tabular}{|c|c|c|c|}
\hline
Method & MNIST classification & IQ classification & Depth regression \\
\hline
& Accuracy & Accuracy & RMSE\\
\hline
CNN (same data) & 0.829 & 0.244 & 0.514\\
MLP (same data) & 0.746 & 0.239 & 0.535\\
CNN (full data) & 0.931 & 0.271 & 0.379 \\
MLP (full data) & 0.927 & 0.261 & 0.443 \\
ChatGPT5 (4-shot prompt) & - & 0.264 & -\\
Gemini2.5pro (4-shot prompt) & - & 0.278 & -\\
PixelFormer \cite{agarwal2023attention} (same data) & - & - & 0.412 \\
PixelFormer \cite{agarwal2023attention} (full data) & - & - & 0.322 \\
\hline
Ours & 0.953 & 0.512 & 0.322 \\
\hline
\end{tabular}
\caption{Results for classification and regression tasks using the proposed framework. `Same data' refers to compared methods that use the same training and adaptation data as the proposed method, whereas `full data' refers to the use of the entire dataset for training for the compared methods.}
\label{tab:class_reg}
\end{table}

\FloatBarrier




\bibliographystyle{alpha}